\documentclass[journal,twocolumn]{IEEEtran}
\usepackage{cite}
\usepackage[english]{babel}

\usepackage{color,soul}
\usepackage{placeins}
\usepackage{capt-of}
\usepackage{booktabs}
\usepackage{float}
\usepackage{amsthm, amsmath, amsfonts}
\usepackage{bbding}
\usepackage{wasysym}
\usepackage{multirow}
\usepackage{graphicx}
\usepackage{pifont}
\usepackage{tabularx}
\usepackage[colorlinks=true, allcolors=blue]{hyperref}

\usepackage{array}
\usepackage{comment}
\usepackage{amssymb}
\usepackage{enumitem}
\usepackage{relsize}
\usepackage[ruled,lined,linesnumbered]{algorithm2e}
\usepackage{xcolor} 
\usepackage{subcaption}
\newcommand*\colourcheck[1]{%
  \expandafter\newcommand\csname #1check\endcsname{\textcolor{#1}{\ding{52}}}%
}
\colourcheck{blue}
\colourcheck{green}
\colourcheck{red}

\newcommand*\colourcross[1]{%
  \expandafter\newcommand\csname #1cross\endcsname{\textcolor{#1}{\ding{55}}}%
}
\colourcross{red}
\colourcross{blue}
\colourcross{green}

\newcolumntype{C}{>{\centering\arraybackslash}X}

\usepackage{upgreek}

\usepackage{filecontents,lipsum}

\theoremstyle{definition}

\title{Empowering Large Language Models in Wireless Communication: A Novel Dataset and Fine-Tuning Framework \vspace{-0.1em} }
\author{Yushen Lin,
        Ruichen Zhang,~\IEEEmembership{Member,~IEEE,}
        Wenqi Huang,~\IEEEmembership{Student Member,~IEEE,}
        Kaidi Wang,~\IEEEmembership{Member,~IEEE,}
        \newline
        Zhiguo Ding,~\IEEEmembership{Fellow,~IEEE,}
        Daniel K. C. So,~\IEEEmembership{Senior Member,~IEEE,}
        and~Dusit Niyato,~\IEEEmembership{Fellow,~IEEE}
\vspace{-2em} 
\thanks{Y. Lin, W. Huang, K. Wang, and Daniel K. C. So are with the School of Electrical and Electronic Engineering, The University of Manchester, M13 9PL, U.K.

Zhiguo Ding is with the Department of Electrical and Electronic Engineering, University of Manchester, Manchester, UK, and the Department of Computer Science, Khalifa University, Abu Dhabi, UAE.

R. Zhang and D. Niyato are with the College of Computing and Data Science, Nanyang Technological University, Singapore.}}

\begin{document}
\maketitle
\begin{abstract}
In this work, we develop a specialized dataset aimed at enhancing the evaluation and fine-tuning of large language models (LLMs) specifically for wireless communication applications. The dataset includes a diverse set of multi-hop questions, including true/false and multiple-choice types, spanning varying difficulty levels from easy to hard. 
By utilizing advanced language models for entity extraction and question generation, rigorous data curation processes are employed to maintain high quality and relevance. Additionally, we introduce a Pointwise V-Information (PVI) based fine-tuning method, providing a detailed theoretical analysis and justification for its use in quantifying the information content of training data with 2.24\% and 1.31\% performance boost for different models compared to baselines, respectively. To demonstrate the effectiveness of the fine-tuned models with the proposed methodologies on practical tasks, we also consider different tasks, including summarizing optimization problems from technical papers and solving the mathematical problems related to non-orthogonal multiple access (NOMA), which are generated by using the proposed multi-agent framework. Simulation results show significant performance gain in summarization tasks with 20.9\% in the ROUGE-L metrics. We also study the scaling laws of fine-tuning LLMs and the challenges LLMs face in the field of wireless communications, offering insights into their adaptation to wireless communication tasks. This dataset and fine-tuning methodology aim to enhance the training and evaluation of LLMs, contributing to advancements in LLMs for wireless communication research and applications.
\end{abstract}


\vspace{-0.2em}
\section{Introduction}

In the rapidly advancing landscape of 6G, artificial intelligence (AI) has emerged as a foundation for introducing innovations in advanced wireless communication technologies \cite{AI_liter1,kaidi_AIFL}. The rapid growth of AI promises unprecedented levels of network efficiency and automation, redefining what these networks can achieve \cite{AI_ref1}. Beyond its impact on wireless communication, AI also drives significant progress in other domains. In particular, large language models (LLMs) have emerged as powerful tools due to their contributions to natural language understanding, knowledge extraction, and decision-making processes \cite{fine_tuneLLM_3}.
Nevertheless, integrating LLMs into wireless communication systems presents significant challenges due to the domain’s inherent complexity \cite{LLM_intro_zhang}. These challenges include solving intricate problems that demand precise, context-aware interpretations of protocols, standards, and dynamic network behaviors essential for optimizing next-generation networks \cite{wireless_LWM}. To maximize the utility of LLMs in wireless communication, advanced datasets and effective fine-tuning strategies are crucial.

Existing wireless communication datasets for LLMs, such as TeleQuAD \cite{dataset_TeleQuAD}, TeleQnA \cite{dataset_teleqna}, and 5GSC \cite{dataset_spec5g}, primarily focus on retrieval-based question-answering and factual recall from standards documents. While these datasets provide a foundation for basic comprehension, they lack the depth or diversity needed to support LLMs for complex reasoning and problem-solving tasks unique in wireless communications. In particular, these datasets fall short in providing multi-hop reasoning, diverse question types, and varying difficulty levels with comprehensive evaluation. As a result, LLMs trained on these datasets are limited in their ability to generalize across different concepts in the domain. Unlocking the full potential of LLMs to revolutionize wireless communications, requires the development of datasets that prioritize both quality and diversity to meet the linguistic and cognitive demands of advanced tasks in this field.

In addition to dataset limitations, fine-tuning LLMs for wireless communication tasks requires methodologies that align with the unique demands of the domain. Fine-tuning involves adapting a pre-trained model to specific tasks by optimizing it with task-relevant data, allowing the model to learn domain-specific knowledge and improve its performance on specialized tasks. Effective fine-tuning enhances model accuracy and ensures computational efficiency, making it suitable for resource-constrained devices. Although prior work \cite{fine_tuneLLM_3} and \cite{fine_tuneLLM_4} have explored fine-tuning for tasks such as physical layer optimization and channel state information (CSI) prediction, existing approaches often fail to address the challenges of scaling to highly technical and dynamic wireless communication scenarios.

To address these limitations, we propose a novel dataset and fine-tuning methodology, specifically designed for wireless communication applications. Our dataset includes a diverse array of multi-hop question types, including true/false and multiple-choice questions, spanning varying difficulty levels from easy to hard and covering a comprehensive range of wireless communication concepts.
In addition to the dataset, we introduce a fine-tuning approach guided by Pointwise V-Information (PVI) metrics inspired by curriculum learning \cite{bengio_learning, PVI}. this approach is applicable to general datasets as it systematically orders dataset instances by difficulty, enabling efficient data selection and utilization during fine-tuning.
By leveraging PVI, the difficulty of each instance within the dataset can be ordered, which can then optimize the selection and utilization of training data, maximizing performance gains while ensuring that the fine-tuned models remain lightweight and suitable for deployment on devices with limited computational resources. 
Through this dataset and fine-tuning methodology, our aim is to bridge the gap between advanced language models and the specialized field of wireless communications, facilitating further research and applications requiring a deep understanding of complex technical concepts.

The dataset and code will be released at: \textit{https://github.com/GTMANChopin/Study-in-Wireless-LLM}. Our contributions are summarized as follows:
\begin{itemize}
\item A comprehensive dataset tailored for wireless communications is created, featuring diverse question types, multi-hop reasoning, and varying complexity levels. This structured dataset not only serves as a robust benchmark for evaluating and fine-tuning large language models on communication-specific reasoning tasks but also offers a versatile resource for domain adaptation and the development of new LLM-based applications in related fields.
\item An effective and robust methodology for automated entity extraction and question generation is implemented, ensuring high technical relevance and quality. Additionally, a rigorous data curation process is introduced to maintain high quality and relevance, facilitating more effective and robust evaluation of LLMs in the wireless communication domain.
\item PVI-based fine-tuning of LLMs is introduced by quantifying the information content learned in wireless communication contexts. Extensive simulation results demonstrate the effectiveness of the proposed dataset and fine-tuning methodology, providing valuable information on model performance and learning dynamics. For further insights, the fine-tuned models are evaluated on two tasks, including summarization and solving mathematical problems.
\item The scaling law of fine-tuning in wireless communications is further analyzed, offering insights into model optimization under different data sizes and computational constraints.
\end{itemize}

The rest of this paper is organized as follows. The literature is reviewed in Section \ref{sec:II}. Section \ref{sec:III} demonstrates the detailed methodologies of data generation. Then, the PVI-based fine-tuning strategy is proposed in Section \ref{sec:IV}. In Section \ref{sec:V}, extensive simulation results are conducted. The scaling laws and challenges for LLMs in wireless communication are studied in Section \ref{sec:scaling}. Finally, we conclude the work in Section \ref{sec:VII}.

\vspace{-0.5em}
\section{Related Works} \label{sec:II}
In this section, we review the literature in areas, including LLMs, datasets, and fine-tuning techniques relevant to wireless communication tasks. We emphasize recent advancements and identify existing gaps that our work aims to address.
\vspace{-0.5em}
\subsection{LLMs in Wireless Communications}
Recent advancements in LLMs have attracted significant interest from the wireless communications community due to their potential to enhance network design, optimization, and management. For example \cite{wirelessLLM_2}, the authors proposed a distributed LLM paradigm tailored for wireless systems, deploying LLMs collaboratively on edge servers and mobile devices. By decomposing the mixture of experts (MoE) layer, the framework leverages parallel capabilities of expert networks on distributed devices, enhancing model performance and reducing end-to-end latency. Further expanding on domain-specific applications, Zhang et al. \cite{wirelessLLM_3} introduced an interactive modeling framework that combines LLMs with retrieval-augmented generation (RAG) techniques to access and apply expert knowledge pertinent to satellite communications. This framework allows LLMs to formulate mathematical models suited to satellite network scenarios, providing real-time adaptability and specialized knowledge handling. Federated learning frameworks, particularly suited for preserving privacy and reducing communication overhead, have been explored for LLM deployment in wireless networks, such as \cite{wirelessLLM_3_FL, wirelessLLM_4_FL}. The framework proposed in \cite{wirelessLLM_3_FL} addresses high processing loads by partitioning the network into client and server sub-networks, where a federated server aggregates client models for updates. In \cite{wirelessLLM_4_FL}, the authors optimized federated learning in wireless communications by introducing personalized federated fine-tuning with low communication overhead, specifically tailored for LLMs in wireless networks, addressing data heterogeneity and client-specific requirements. Furthermore, \cite{wirelessLLM_5} employed LLM-based combinatorial optimization algorithms to determine the number and placement of wireless access points; thus improving network performance. 
Despite these advancements, significant challenges remain, such as adapting LLMs to the resource constraints of wireless devices and enabling multi-hop reasoning for complex problem-solving scenarios.
\vspace{-0.3em}
\subsection{Domain-Specific Datasets}
Specialized datasets play a critical role in the evaluation and refinement of LLMs for wireless communication applications. For example, the TeleQuAD dataset \cite{dataset_TeleQuAD}, contains 2,021 question-answer (QA) pairs extracted from 3GPP standards, designed to test LLMs on telecom-related questions. TeleQnA \cite{dataset_teleqna} provided 10,000 multiple-choice questions across categories such as lexicon, research overview, research publications, standards overview, and standards specifications. 
Similarly, \cite{dataset_standard} introduced a dataset of 2400 QA pairs based on the 3GPP and IEEE specifications. The 5G Standards Corpus (5GSC) \cite{dataset_spec5g} contained 2401 QA pairs related to 5G standards, facilitating the evaluation of the comprehension of the models for 5G technologies and protocols. 

Although datasets are essential for understanding and evaluating model performance, assessing and studying the difficulty of individual instances in LLM tasks is also crucial for guiding curriculum learning strategies. There are several studies on determining the difficulty of each instance in the dataset, such as \cite{dataset_diff_curriculum,dataset_diff_CDA,dataset_diff_squad,dataset_diff_IRT}. For example, the paper in\cite{dataset_diff_curriculum} introduced a curriculum-based method, which systematically escalates difficulty based on educational levels and cognitive complexity. In \cite{dataset_diff_CDA}, the authors proposed a competition-based model to rank the difficulty of the question using pairwise comparisons of users and questions on forums. In \cite{dataset_diff_squad}, the authors investigated how linguistic features such as syntax complexity and dependency structures affect the difficulty of the question. The authors in \cite{dataset_diff_IRT} applied item response theory, a psychometric tool, to measure the difficulty of the AI classification task by treating instances as ‘items’ and classifiers as ‘respondents’.

\vspace{-0.3em}
\subsection{Wireless Fine-Tuning}
Fine-tuning LLMs for wireless communication tasks has emerged as a key research area, aiming to customize these models for domain-specific challenges. 
In \cite{fine_tuneLLM_3}, BART was fine-tuned to integrate into AI communication systems on devices to improve its ability to handle physical layer communication challenges, such as noise robustness and efficient data compression, while ensuring generalization across diverse and unseen scenarios. In the study \cite{fine_tune_telegpt}, LLMs were fine-tuned to effectively follow telecom-specific instructions using supervised fine-tuning (SFT) with a custom dataset of telecom-related tasks. The authors \cite{fine_tuneLLM_4} proposed fine-tuning a pre-trained GPT-2 model to predict future downlink CSI sequences using historical uplink CSI data. By fine-tuning, this approach leveraged the modeling and generalization strengths of LLMs to improve prediction accuracy.

Although these works demonstrate promising advances, they leave certain aspects unaddressed. 
Notably, understanding the challenges facing LLMs in wireless communications remains a critical area of investigation. Furthermore, enabling effective multi-hop reasoning within wireless communication tasks continues to be an open challenge, constraining the models' ability to address complex, multi-layered problem-solving scenarios. Bridging these gaps is essential for advancing the integration of LLMs into wireless communication systems, facilitating the development of more robust and efficient network solutions.

\vspace{-0.2em}
\section{Data Generation Methodology} \label{sec:III}

In this section, we present the methodology employed to construct a comprehensive and high-quality dataset tailored for wireless communications. The process involves four key components, i.e., data source retrieval, entity generation, data curation, and example construction. We use uplink non-orthogonal multiple access (NOMA) as an example to demonstrate the data generation process. To elucidate the technical foundation of NOMA, consider a scenario in which two users are transmitted via sub-channel $i$.
The achievable data rate for the first decoded user can be expressed as follows:
\begin{equation}
R_{i,1}^{\textrm{noma}} = B \log_2\left(1 + \frac{p_{i,1} |h_{i,1}|^2}{p_{i,2} |h_{i,2}|^2 + 1}\right), 
\end{equation}
where $B$ is the bandwidth, $|h_{i,1}|^2$ and $|h_{i,2}|^2$ represent the normalized channel gain of the first and second users, respectively \cite{kaidi}, in sub-channel $i$, $p_{i,1}$ and $p_{i,2}$ denote the transmit power allocated to the respective users. It is assumed that $|h_{i,1}|^2 \ge |h_{i,2}|^2$. After the first user's signal is successfully decoded and removed through successive interference cancellation (SIC), the second user’s signal can be detected without interference. Consequently, the achievable data rate for the second decoded user is:
\begin{equation}
R_{i,2}^{\textrm{noma}} = B \log_2\left(1 + p_{i,2} |h_{i,2}|^2\right). 
\end{equation}

\begin{figure}[t!]
    \vspace{-2mm}
    \centering
    \includegraphics[width=1\linewidth]{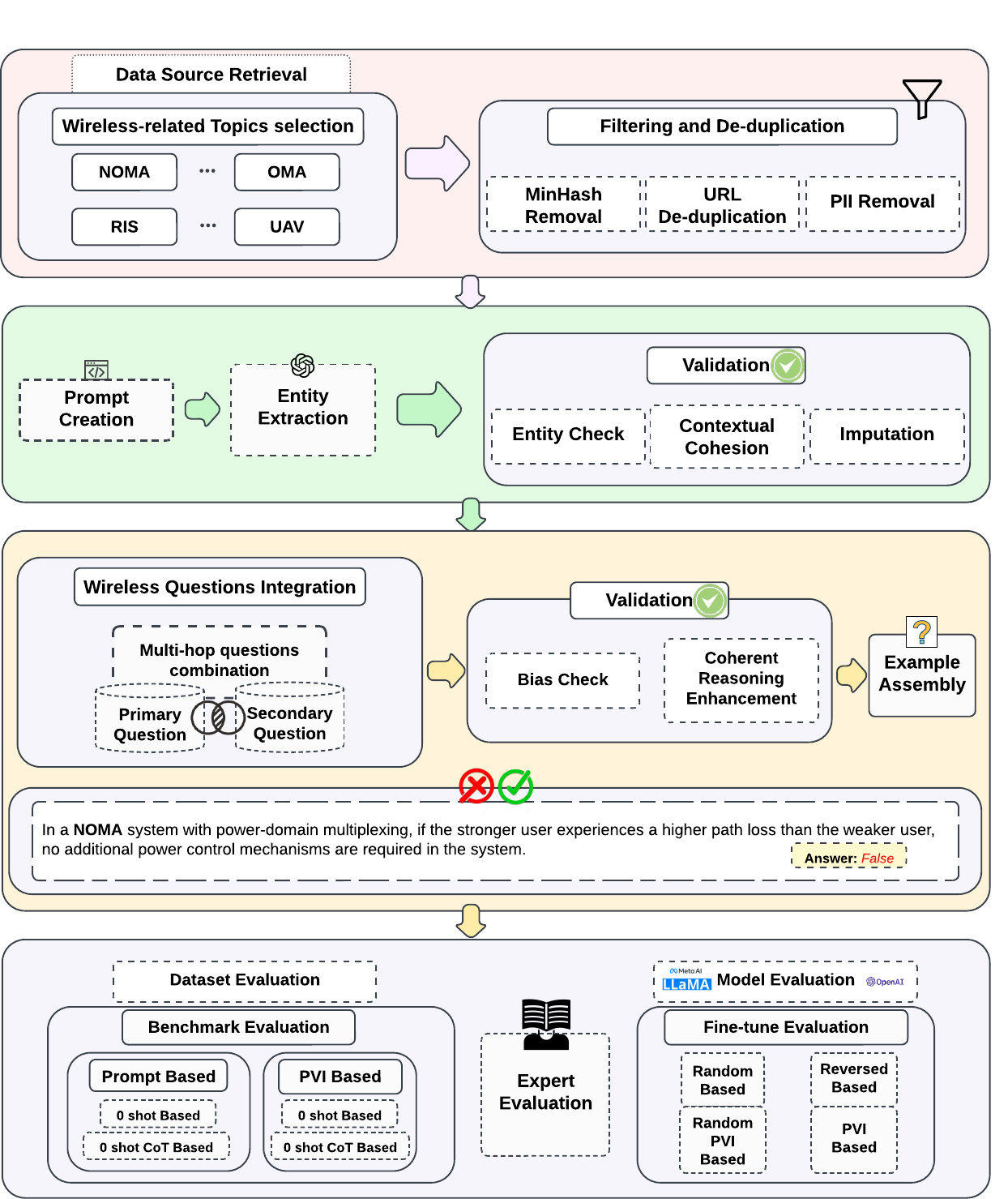}
    \caption{Structure and outline of Section \ref{sec:III} and Section \ref{sec:IV}. The methodology outlines constructing a high-quality wireless communications dataset by retrieving and sanitizing articles, automatically extracting key technical entities with LLMs, and curating coherent multi-hop reasoning examples. Using NOMA as an example, the process integrates sequential subquestions into complex queries, ensures logical consistency through reasoning chains, validates answers, and applies bias mitigation strategies to maintain accuracy and impartiality.}
    \label{fig:structure}
    \vspace{-2mm}
\end{figure}

\vspace*{-0.5cm}
\begin{table*}[t]
\vspace{-2mm}
\centering
\renewcommand{\arraystretch}{1.5} 
\setlength{\tabcolsep}{6pt}
\begin{tabular}{|p{0.75\textwidth}|p{0.11\textwidth}|}
\hline

\textbf{Question and Reasoning Types with Example(s)} & \textbf{Question Type} \\ \hline

\textbf{Example of constructions of multiple-choice questions} \newline
\textbf{Context A:} \textcolor{red}{NOMA} is the technique that \setulcolor{green}
\ul{allows multiple users to share the same time and frequency} ... \newline
\textbf{\textcolor{green}{$q_{1,n}$}:}
Which wireless communication technique allows multiple users to share time and frequency resources? \newline
\textbf{Context B:} In \textcolor{red}{NOMA} system, \setulcolor{orange}
\ul{power control plays a critical role in separating the signals} ... \newline
\textbf{\textcolor{orange}{$s_{2,n}$}:} 
Which wireless communication technique allocates different power to users based on the channel conditions? \newline
\textbf{Integrated Q from \textcolor{green}{$q_{1,n}$} and \textcolor{orange}{$s_{2,n}$}:} Which telecommunications technique improves spectral efficiency by allowing multiple users to share the same time and frequency resources and potentially allocating different power among those users based on channel conditions? \newline
A. \textcolor{red}{NOMA} \quad B. TDMA \quad C. CDMA \quad D. OFDM 
& \textbf{Multiple Choice} \\ \hline

\textbf{Example of constructions of True/False questions} \newline
\textbf{Context A:} The available \textcolor{red}{NOMA} techniques can broadly be divided into \setulcolor{green}\ul{two major categories, i.e., power-domain {\color{red}{NOMA}} and code-domain {\color{red}{NOMA}} } ... \newline
\textbf{\textcolor{green}{$q_{1,n}$}:} Power-domain \textcolor{red}{NOMA} and code-domain \textcolor{red}{NOMA} are two major categories of NOMA techniques. \greencheck \newline
\textbf{Context B:} In \textcolor{red}{NOMA} introduces \setulcolor{orange}\ul{additional interference by allowing users to be superimposed on the same resource} ... \newline
\textbf{\textcolor{orange}{$s_{2,n}$}:} \textcolor{red}{NOMA} eliminates interference by assigning unique resources to each user. \redcross \newline
\textbf{Integrated Q from \textcolor{green}{$q_{1,n}$} and \textcolor{orange}{$s_{2,n}$}:} Power-domain \textcolor{red}{NOMA} eliminates the interference by assigning different optimized power to users. \greencheck
& \textbf{True/False} \\ \hline

\end{tabular}
\caption{The illustrations of NOMA-related examples of construction of different types of questions, i.e., multiple-choice, true/false, hard reasoning questions. \textcolor{red}{Red} represents the entity, underlines in \textcolor{green}{green} and \textcolor{orange}{orange} represent the meaningful facts from the context A and B that are directly related to the questions, respectively. }
\vspace{-2mm}
\end{table*}

\subsection{Data Source Retrieval}
The dataset construction begins with the identification and extraction of articles on key topics of wireless communication, denoted by $\textbf{T}$, i.e., NOMA. 
The topics guide the search queries via the MediaWiki APIs \cite{MediaWikiAPI}. To eliminate redundancy, repeated documents are removed. Filtered by retaining only the most recent context for each URL using the MinHash algorithm \cite{miniHash}. Additionally, all personally identifiable information (PII) is removed from the dataset. 
Specifically, given a context $X'$, PII subsets $p \in X'$ are identified and removed, resulting in a sanitized context $X=X'$. For example, ‘User 1 and User 2 share the same resources in a NOMA system with different power levels’.

\subsection{Entity Generation}
Key technical terms, referred to as ‘entities’, are automatically extracted from sanitized contexts using LLMs. Each entity represents a core concept within wireless communications, such as power allocation in NOMA. 

Custom prompts are constructed to extract the entities from each context are constructed, i.e., 
the prompt used in entity extraction (see Appendix C). For the $n$-th context $x_n$, the prompt template $x_n'$ is formulated as follows:
\begin{equation}\label{eq:prompt}
x_n' = \text{“[$x_n$]. In the given context, the primary entity is [$e_n$].”}
\end{equation}
where $x_n'$ denotes the prompt template with placeholders for input $x_n$ and entity $e_n$, where $e_n \in \mathbf{E}$, and $\mathbf{E}$ denotes the set of key entities to be extracted from the context $x_n$.

The output $n$-th entity that maximizes the likelihood of the filled prompt can be formulated as follows:
\begin{equation}\label{eq:extract_entity}
e_n = \arg \max P\left(l_{\text{f}}(x_n', e_n'); \mathbf{w}\right),
\end{equation}
where $l_{\text{f}}(x_n', e_n')$ refers to the filled prompt with a candidate entity $e_n'$, and $\mathbf{w}$ represents the model parameters.
Extracted entities are further validated to ensure that they are non-empty, contextually accurate, and aligned with the topic.
\vspace{-0.2em}
\subsection{Data Curation and Example Assembly}
A rigorous curation process is performed to ensure the high quality and relevance of the generated examples. Each context and its corresponding questions are accessed based on criteria such as length, relevance, and alignment with the original context.
For example, an entity $e_n$ must be present and meaningfully integrated within its associated context $x_n$; thus maintaining alignment and relevance to the task, i.e., $e_n \in x_n$.

An essential component of this process is the generation and integration of questions to simulate multi-hop reasoning. Multi-hop reasoning requires integrating multiple pieces of evidence from different contexts to answer a question. This is more challenging than single-hop reasoning and is critical for tasks that require deeper understanding, such as answering complex queries or solving problems that mimic real-world scenarios \cite{multihop_ref}. 
The question generation process aims to produce two or more related questions for each question-answer pair. Initially, a primary question ($q_{1,n}$) is generated based on the content of the article, where the answer is typically an entity. The secondary question ($s_{2,n}$) is then derived from the additional context that has been previously extracted. These questions are integrated using LLMs with designed prompts (see Appendix C) to maintain relevance and coherence.

To create a coherent multi-hop question, $q_{1,n}$ and $s_{2,n}$ are combined into an integrated question $q_n$. The integration process involves constructing a prompt that unifies $q_{1,n}$ and $s_{2,n}$ into a single question, i.e., in the process described in Table \ref{fig:structure}. The prompt is designed to link the reasoning steps in a manner that necessitates the integration of information from both questions (see Appendix C). Utilizing LLMs, the model infers the final integrated question, capturing the complexity of reasoning required to answer both subquestions. The model aims to generate the most probable $n$-th integrated question $q_n$ that can be expressed as Eq. \eqref{eq:integrate} in Algorithm \ref{alg:dataset}, in which $q'$ represents possible integrated questions given the subquestions, $\mathbf{Q}'$ denotes the set of all possible integrated questions, $x_{g}'$ is the prompt template including the context of both subquestions and any relevant background information, and $l_{\text{f}}(x_g', q_{1,n}, s_{2,n}, q')$ is the filled prompt that integrates the subquestions into the final question $q_n$.

Ensuring the logical coherence of reasoning chains in multi-hop questions is critical. A reasoning chain ${r_1, r_2, \ldots, r_i}$ is considered valid if it logically progresses to a final answer $\alpha$:
\begin{equation}\label{eq:chain}
r_1 \rightarrow r_2 \rightarrow \cdots \rightarrow r_i \rightarrow \alpha, 
\end{equation}
where $r_i$ denotes an intermediate reasoning step. If gaps or inconsistencies are detected, an imputation process will regenerate missing elements, such as omitted portions of the context or questions.

\begin{algorithm}[h]
\caption{Data Generation Methodology}
\label{alg:dataset}

\DontPrintSemicolon 
\KwIn{Topics list $\mathbf{T}$: Core wireless communication topics, Pre-trained LLM with parameters $\mathbf{w}$.}
\KwOut{Dataset $\mathcal{D}$.}

\For{each topic $t \in \mathbf{T}$}{
    Retrieve articles $\mathbf{A}_t$\;
}
Collect all articles: $\mathbf{A} \leftarrow \bigcup_{t} \mathbf{A}_t$\;
Remove duplicates using MinHash and sanitize contexts by removing PII to get $\mathbf{X}$\;

\For{each context $x_n \in \mathbf{X}$}{
    Construct prompt with Eq. \eqref{eq:prompt}

    \If{$e_n \neq \emptyset$ \textbf{and} $e_n \in x_n$}{
        Generate $q_{1,n}$ and $s_{2,n}$\;
        Integrate into multi-hop question:
        \begin{equation}\label{eq:integrate}
        q_n = \arg \max_{q' \in \mathbf{Q}'} P\left(l_{\text{f}}(x_g', q_{1,n}, q_{2,n}, q'); \mathbf{w}\right)
        \end{equation}
        Derive answer and explanation of Eq. \eqref{eq:answer}\;
        Ensure reasoning chain of Eq. \eqref{eq:chain} is valid\;
        \If{bias detected in $y_n$}{
            Apply bias mitigation strategies\;
        }
        Add $(q_n, y_n)$ to $\mathcal{D}$\;
    }
}
\end{algorithm}

Each assembled example comprises several components: the integrated multi-hop question $q_n$, the answer $\alpha_n$, the individual subquestions ($q_{1,n}$ and $s_{2,n}$), the extracted entity $e_n$, explanations of the answer, and the original article text. These components collectively form a structured example, facilitating the multi-hop reasoning process and illustrating how complex queries can be deconstructed and solved. The probabilistic reasoning process of LLMs in deriving the answer for $q_n$ can be expressed as follows:

\begin{equation}\label{eq:answer}
\alpha_n = \arg\max_{\alpha' \in \mathbf{A}'} P\left(\alpha' \mid q_n; \mathbf{w}\right), 
\end{equation}
where $\alpha'$ denotes possible answers to question $q_n$, and $\mathbf{A}'$ is the set of possible answers.

After completion of these processes, the dataset $\mathcal{D}$ becomes available for evaluation, where each instance consists of a pair of question-answer questions ($q_n$, $\alpha_n$). For the fine-tuning task in our work, each instance includes a question-answer-explanation triplet ($q_n$, $y_n$), represented as $\mathbf{S} = {(q_1, y_1), (q_2, y_2), \ldots, (q_n, y_n)}$, where $\mathbf{S} \subset \mathcal{D}$ and $y_n$ denote the explanation accompanying the answer of the $n$-th question $q_n$. The difficulty of each instance in $\mathcal{D}$ is ordered based on the PVI, which is described in the following section.

To mitigate the biases inherent in LLMs, we implemented strategies such as the Quiet-STaR prompt \cite{Quiet-STaR} to identify and address biased content. Furthermore, domain experts review the dataset to ensure accuracy and impartiality, minimizing the risk of propagating biased knowledge (see Appendix C).

\section{Proposed PVI-based Fine-Tuning}\label{sec:IV}
\subsection{Parameter-Efficient Fine-Tuning}

In practical scenarios, the deployment and fine-tuning of LLMs on devices with limited computational resources, such as internet-of-things (IoT) devices, presents significant challenges. Full fine-tuning of LLMs demands substantial computational power and memory, which is impractical for constrained devices. For example, fully fine-tuning a relatively small language model with 8 billion parameters using the widely adopted AdamW optimizer \cite{adamW}, requires at least 59.83 GB of memory for optimizer states and 54.08 GB for activations. These requirements far exceed the capabilities of most edge devices, highlighting the need for alternative fine-tuning approaches that minimize computational overhead. To address this issue, we employ parameter-efficient fine-tuning (PEFT) methods that adapt the model to specific tasks without the computational cost of full model updates.

Among these methods, low-rank adaptation (LoRA) is widely adopted for its efficiency and effectiveness \cite{LoRA}. LoRA utilizes two small trainable matrices, i.e., $\mathbf{A} \in \mathbb{R}^{m \times r}$ and $\mathbf{B} \in \mathbb{R}^{r \times n}$ to update the original weight matrix with $r \ll \min(m,n)$. Here, $r$ denotes the rank of the low-rank decomposition, $m$ denotes the input dimension of the weight matrix, and $n$ is the output dimension of the original weight matrix. During fine-tuning, only the matrices $\mathbf{A}$ and $\mathbf{B}$ are updated, while the large weight matrix $\mathbf{W}$ remains unchanged, significantly reducing memory and computational requirements.

Our fine-tuning process leverages the training dataset comprising questions-answers-explanations pairs $\mathbf{S}$. The optimization objective for fine-tuning LLM is formulated as:\footnote{For simplicity of notations, we drop subscript from here for clear demonstration.}
\begin{equation}\label{eq:LoRA}
\max_{\mathbf{\Xi}} \sum_{(q, y) \in \mathbf{S}} \sum_{t=1}^{|y|} \log \left(p_{\mathbf{w}_0 + \mathbf{\Xi}}(y_t | q, y_{<t})\right),
\end{equation}
where $\mathbf{\Xi}$ denotes the low-rank decomposition parameters. The inner summation iterates all the tokens in the explanations, computing the total log-likelihood of generating the explanation $y$ token by token, conditioned on the question $q$ and the previous tokens $y_{<t}$.

\begin{figure}[h]
    \vspace{-2mm}
    \centering
    \includegraphics[width=\linewidth]{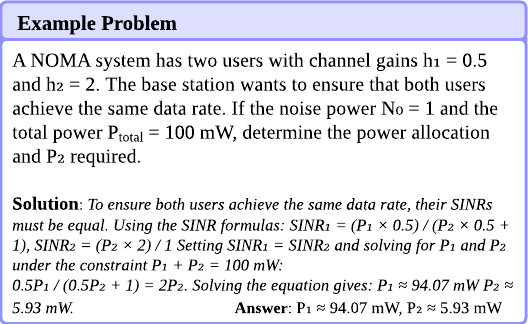}
    \caption{Example question generated by using multi-agent (See Appendix D).}
    \vspace{-4mm}
    \label{fig:q2}
    \vspace{-2mm}
\end{figure}

\vspace{-0.5em}
\subsection{Pointwise V-Information}
Traditional fine-tuning approaches often fail to quantify the meaningful information a model extracts from the training data. This limitation hinders the optimization of the fine-tuning process, especially under computational constraints or when maximizing learning efficiency is critical.
PVI provides a theoretical framework to measure the amount of ‘usable' information a model can extract from individual instances, distinguishing itself from traditional metrics such as Shannon's information \cite{PVI}. Unlike Shannon information, which measures average information content, PVI focuses on the information that is utilized by the model within its computational constraints.

PVI quantifies the additional information that a model gains when presented with specific input data. Given a predictive family $\mathcal{V}$ \footnote{Predictive family is a subset of all possible mappings from sample spaces to the set of all possible probability distributions over the label space (possible outcomes). See reference \cite{PVI}.}, let $m[\emptyset](y)$ denote the probability of predicting $y$ without access to the input $q$, i.e.,
\begin{equation}
    m[\emptyset](y) = p(y),
\end{equation}
where $\emptyset$ denotes null input providing no contextual information regarding $q$. In our fine-tuning process, it can be set to an empty string.
When the model gains access to the question $q$, its probability of predicting $y$ is expressed as follows:
\begin{equation}
    m'[q](y) = p(y \mid q),
\end{equation}
where $m$ and $m'$ correspond to the models without and with access to the input $q$, respectively, and $m, m' \in V$.  For instance, if $\mathcal{V}$ represents the GPT or LLaMA family, $m$ and $m'$ correspond to these models fine-tuned without and with input data.
\subsection{PVI for an Instance $(q, y)$}
PVI measures the usable information usable by a model for specific instances $(q, h)$ \cite{PVI}. Based on our dataset $\mathcal{D}$, the PVI is defined as:

\begin{equation}
    \begin{aligned}
    & \text{PVI}(q \to y) = -\log_2 m[\emptyset](y) + \log_2 m'[q](y) \\
    & \quad \quad \quad \quad \quad = -\log_2 p(y) + \log_2 p(y \mid q),
    \end{aligned}  
\end{equation}
which further simplifies to:
\begin{equation} \label{eq: PVI}
    \text{PVI}(q \to y) = \log_2 \frac{p(y \mid q)}{p(y)}.
\end{equation}
Here the $\log_2$ is used to measure the entropy in bits of information. This expression quantifies the additional information about $y$ accessible when the question $q$ is provided. For example, instances with higher PVI are simpler for $\mathcal{V}$ to handle, in which a greater PVI increases the likelihood of accurate prediction.

\begin{figure*}[t!]
    \vspace{-1mm}  
    \includegraphics[width=\textwidth]{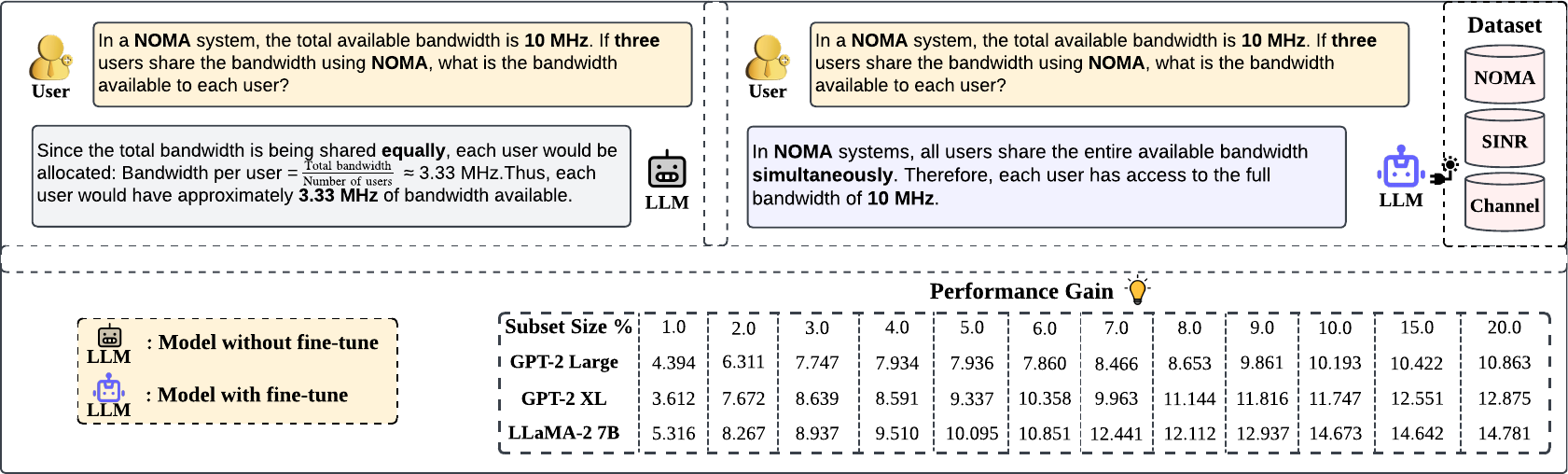}
    \captionof{figure}{Performance gain comparison across subset sizes for GPT-2 Large, GPT-2 XL, and LLaMA-2 7B models. While fine-tuning leads to consistent performance improvements, emphasizing its advantage on task-specific enhancements. Interestingly, the relatively straightforward questions, exemplified by the one illustrated in this figure, were evaluated across various LLMs, with even several advanced models failing to produce the correct answers, including LLaMA-3.1 8B \cite{llama3}, GPT-4o-mini, etc.}
    \label{fig:performance_example}
\end{figure*}

To evaluate PVI at the token level, we express $y$ as a sequence $y_1, y_2, \dots, y_{|y|}$. The probabilities can be decomposed as follows: 
\begin{equation} p(y) = \prod_{t=1}^{|y|} p(y_t \mid y_{<t}), \quad p(y|q) = \prod_{t=1}^{|y|} p(y_t \mid q, y_{<t}). \end{equation} Substituting these into the PVI definition:
\begin{align} \text{PVI}(q \to y) &= \sum_{t=1}^{|y|} \left(\log_2 p(y_t \mid q, y_{<t}) - \log_2 p(y_t \mid y_{<t}) \right). \end{align}

This equation computes the per-token variable importance by summing the difference in log-probabilities for each token when conditioned on $q$ versus when not conditioned on $q$.
The probabilities conditioned on the model parameters become:
\begin{equation}
    p(y_t \mid y_{<t}) = p_{\mathbf{w}_0 + \mathbf{\Xi}}(y_t \mid y_{<t}),
\end{equation}

\begin{equation}
    p(y_t \mid q, y_{<t}) = p_{\mathbf{w}_0 + \mathbf{\Xi}}(y_t \mid q, y_{<t}).
\end{equation}

The LoRA parameters $\mathbf{\Xi}$ incorporate low-rank adaptations of the weights of the base model $\mathbf{w}_0$. By substituting these parameterized probabilities into the PVI expression, we directly measure how the adapted model (with minimal additional parameters) benefits from the input $q$. In other words, we quantify how much domain-specific value the LoRA-based fine-tuning extracts from each training instance, providing a token-level understanding of the model’s enhanced predictive capability under computational constraints.
Thus, the PVI for the instance $(q, y)$ in the context of fine-tuning by using LoRA can be expressed as follows:

\begin{equation}
    \begin{aligned}
       & \text{PVI}(q \to y) = \sum_{t=1}^{|y|} \Bigl( \log_2 p_{\mathbf{w}_0 + \mathbf{\Xi}}(y_t \mid q, y_{<t}) \\
       &  \quad \quad \quad \quad \quad \quad  - \log_2 p_{\mathbf{w}_0 + \mathbf{\Xi}}(y_t \mid y_{<t})\Bigl),
    \end{aligned}
\end{equation}
which can be expressed as:
\begin{equation}
    \text{PVI}(q \to y) = \log_2 \frac{p_{\mathbf{w}_0 + \mathbf{\Xi}}(y \mid q)}{p_{\mathbf{w}_0 + \mathbf{\Xi}}(y)}.
\end{equation}

\section{Simulation}\label{sec:V}

\begin{table*}[t!]
    \centering
    \caption{Accuracy of different models on the dataset using zero-shot and zero-shot CoT.}
    \label{tab:accuracy_table}
    \small 
    \setlength{\tabcolsep}{4pt} 
    \resizebox{\textwidth}{!}{ 
    \begin{tabular}{l@{\hspace{4pt}}cccccccc}
        \toprule
        \textbf{Models} & \textbf{GPT-2 XL} & \textbf{LLaMA 1.1B} & \textbf{GPT-2 XL} & \textbf{LLaMA 2 7B} & \textbf{LLaMA 2 13B} & \textbf{GPT-3.5 Turbo}\footnotemark & \textbf{GPT4o-mini} & \textbf{GPT4o} \\
        \midrule
        \textbf{Size} & 774M & 1.1B & 1.5B & 6.7B & 13B & - & - & - \\
        \addlinespace
        \cmidrule(lr){2-9}
        \textbf{Zero-shot}(overall) & 0.190 & 0.378 & 0.336 & 0.124 & 0.155 & 0.353 & 0.641 & 0.664 \\
        \textbf{Zero-shot CoT}(overall) & 0.162 & 0.390 & 0.264 & 0.237 & 0.254 & 0.365 & 0.647 & 0.686 \\
        \midrule
        \textbf{Easy Zero-shot} & 0.193 & 0.429 & 0.344 & 0.131 & 0.151 & 0.536 & 0.759 & 0.807 \\
        \textbf{Easy CoT} & 0.165 & 0.430 & 0.271 & 0.234 & 0.276 & 0.582 & 0.761 & 0.827 \\
        \midrule
        \textbf{Medium Zero-shot} & 0.191 & 0.391 & 0.337 & 0.123 & 0.156 & 0.332 & 0.646 & 0.661 \\
        \textbf{Medium CoT} & 0.164 & 0.404 & 0.266 & 0.242 & 0.252 & 0.320 & 0.654 & 0.670 \\
        \midrule
        \textbf{Hard Zero-shot} & 0.187 & 0.313 & 0.327 & 0.120 & 0.159 & 0.192 & 0.517 & 0.523 \\
        \textbf{Hard CoT} & 0.158 & 0.336 & 0.255 & 0.236 & 0.236 & 0.193 & 0.525 & 0.562 \\
        \bottomrule
    \end{tabular}
    }
    \vspace{-1em} 
\end{table*}

In this section, extensive simulations and evaluations are presented to evaluate and demonstrate the effectiveness of our proposed dataset and the fine-tuning methodology.

\textbf{Data Generation} The dataset is constructed using advanced LLMs for entity extraction and question generation. Entities are extracted using GPT-4o-mini, while GPT-4o (gpt-4o-2024-08-06) is employed for integrating subquestions into multi-hop questions. The choice balances cost and performance for entity extraction and ensures quality for question integration.

\textbf{Dataset Evaluation} To evaluate the dataset, experiments are conducted using various LLMs, including models from the GPT and LLaMA families. We employ zero-shot and zero-shot Chain-of-Thought (CoT) prompt strategies \cite{cot}. The tasks in our experiments have varying difficulty levels, which are ordered based on PVI as demonstrated in Section \ref{sec:IV}.

\textbf{Training Details} Fine-tuning experiments, including the investigation of scaling laws in the context of wireless communication shown in Fig. \ref{fig:scaling_alpha}, are conducted on one NVIDIA A100 PCIe 80 GB. Models exceeding one billion parameters are fine-tuned with a learning rate of $5e-4$, while smaller models use $5e-5$ \cite{fine_tune_learning_rate}. The LLaMA 2 7B model is fine-tuned using LoRA \cite{LoRA} with rank $r=8$, employing the AdamW optimizer ($\beta_1=0.9$, $\beta_2=0.999$) and a weight decay of 0.1. The maximum token limit is set to 256, with a batch size of 16 to manage memory constraints. Furthermore, all models are trained for 3 epochs \cite{fine_tune_literature2}. The dataset is split into 80\% for training and 20\% for testing, ensuring that the test set contains examples of varying difficulty levels. The difficulty of each instance $\mathcal{D}$ is ordered by PVI from low to high and then clustered into three levels: easy, medium and hard using the clustering method of K-means.

\subsection{Dataset Evaluations}
Table \ref{tab:accuracy_table} compares the overall accuracy of the models using zero-shot and zero-shot CoT strategies. Comparison between different model sizes shows the base accuracy achieved by each. Smaller and mid-sized models show limited benefits from CoT prompting, indicating that they lack the capacity to utilize structured reasoning for complex domain-specific queries in wireless communications. Additionally, complex wireless communication queries, which often require multi-step reasoning or a deep understanding of technical concepts, remain challenging even for advanced models. For instance, it can be seen that the accuracies are relatively low even for the advanced models when it comes to hard multi-hop questions (refer to Section \ref{sec:IV}). 
While CoT improves performance for larger models such as GPT-4o and GPT-3.5 Turbo on medium and hard questions, smaller models are unable to perform the nuanced reasoning essential for handling such intricate domain-specific queries. However, relying solely on large models is often impractical due to their high computational costs and resource demands, especially for local deployments. 
\footnotetext{GPT-3.5-turbo represents an enhancement over text-davinci-003 \cite{openai_model_index}, and is presumed to have 175 billion parameters. However, OpenAI has not officially confirmed this information in any public sources.}Table \ref{tab:accuracy_table} illustrates the performance comparison of language models across various parameter scales, from OPT-350M to GPT-4o, utilizing both zero-shot and zero-shot CoT methodologies based on different difficulty levels. A clear positive correlation exists between model size and accuracy, with larger models (e.g. GPT-4o, GPT-4o-mini) substantially outperforming smaller ones (e.g. OPT-350M, GPT2-large) except tinyLLaMA 1.1B \cite{tinyllama}. Accuracy consistently decreases across almost all models as task difficulty increases from easy to hard. The performance gap is most evident in larger models for more complex tasks. 
The benefits of CoT reasoning are minimal overall, with only slight improvements observed in larger models (especially the GPT-4o series) on harder tasks. Additionally, a noticeable performance improvement occurs between relatively large LLaMA models (for 7 and 13B parameters) and the smaller GPT series. We conclude, based on our observations during the simulation, that the models of the LLaMA 2 family have minimal capability to follow instructions.
To seek the reason, we tune the parameter of the maximum number of tokens to generate and the prompt used in the LLaMA family. We find that the models from the LLaMA family struggle to follow instructions in many formats and styles (e.g., answering within a given length, using specific letter cases, etc.). The results provided in Table \ref{tab:accuracy_table} are based on 30 tokens. That is, if the answer to a question (true/false or A, B, C, D) is not given within 30 tokens, then the answers are marked incorrect. 

\begin{figure}[htp]
    \centering
    \includegraphics[width=1\linewidth]{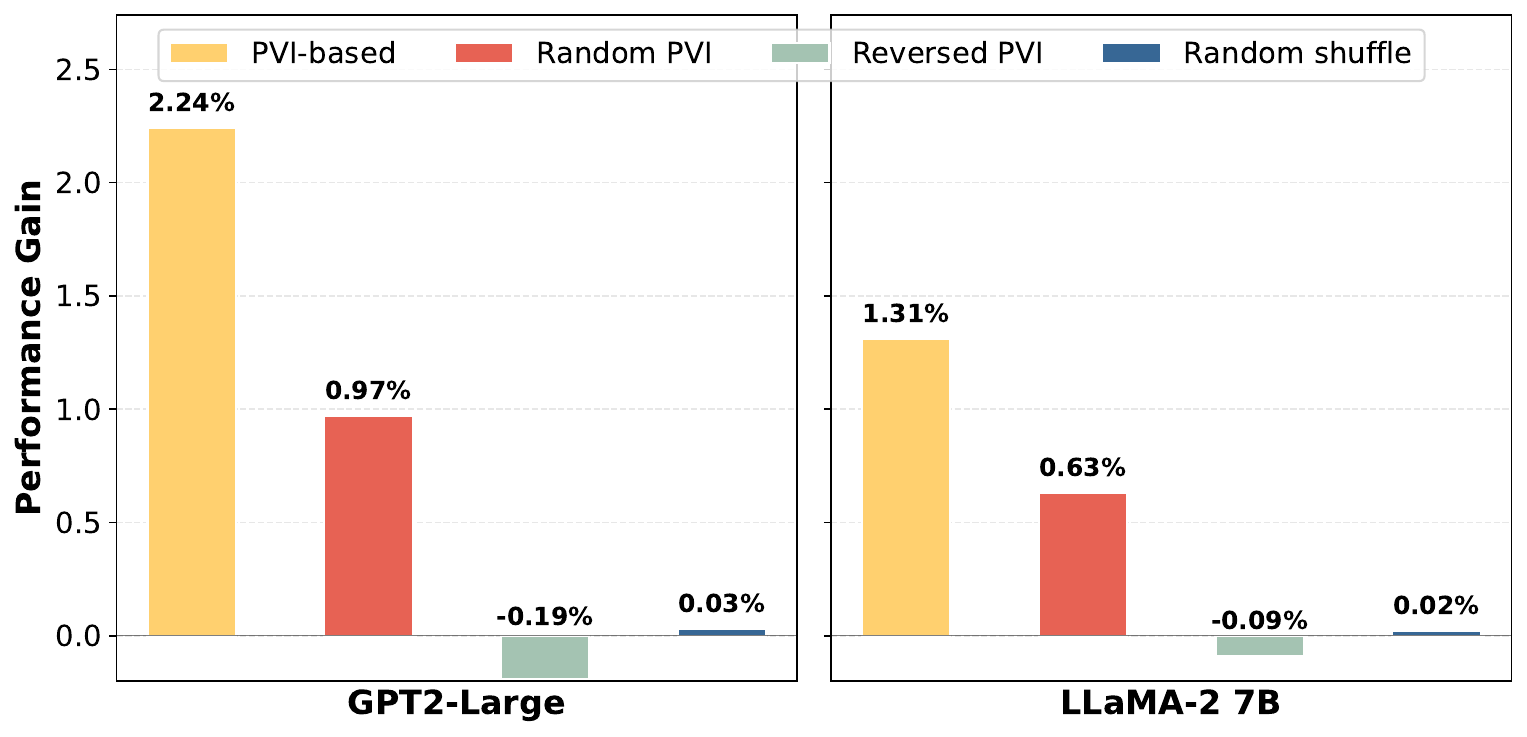}
    \caption{Comparsions of performance gains across different data ordering strategies for GPT2-large and LLaMA-2 7B. }
    \label{fig:easy_2_hard}
\end{figure}
\subsection{Fine-tuning Evaluations}
To demonstrate the performance gain through the PVI-based fine-tuning strategy, we compare our approach with the strategies listed below \cite{PVI}.

\begin{itemize} 
\item \textbf{Standard Fine-Tuning (Random Shuffle)}: The model is fine-tuned on the entire dataset without any ordering.
\item \textbf{Random PVI}: Each instance in $\mathcal{D}$ is randomly shuffled within each difficulty level group prior to fine-tuning. 

\item \textbf{Reverse PVI}: Each instance in $\mathcal{D}$ is ordered from hardest to easiest based on PVI values.
\end{itemize}
We evaluate the performance gains achieved by fine-tuning three language models—GPT-2 Large, GPT-2 XL, and LLaMA-2 7B—across varying subset sizes of the training data in Fig. \ref{fig:performance_example}. All models show consistent improvement as the subset size increases, highlighting the importance of data availability in fine-tuning. LLaMA-2 7B consistently outperforms GPT-2 Large and GPT-2 XL, achieving higher performance gains at each subset size, particularly noticeable at larger subsets (e.g., 10\% and 20\%). GPT-2 XL exhibits better performance than GPT-2 Large in most subset sizes, showcasing the benefits of increased model capacity in wireless communications Q\&A. Although performance improves with larger subsets, the rate of gain decreases, indicating diminishing returns on performance improvements with additional data. 

Fig. \ref{fig:easy_2_hard} shows that PVI-based fine-tuning shows the strongest positive impact (2.24 for GPT2-large, 1.31 for LLaMA-2), while random PVI demonstrates measurable gains (0.97 and 0.63 respectively) compared to other baselines. In contrast, reversed PVI and random shuffle methods show minimal or slightly negative effects, suggesting that strategic data ordering significantly influences model performance as discussed in Section \ref{sec:IV}. These results indicate that computationally constrained wireless communications users can further benefit from the proposed method by using PVI-based ordering to optimize fine-tuning efficiency, achieving enhanced performance with limited computational resources. It is worth noting that this strategy remains feasible for users with sufficient capability to perform standard fine-tuning.

\begin{figure}[htp]
    \centering
    \begin{subfigure}[htp]{0.49\linewidth}
        \centering
        \includegraphics[width=0.99\linewidth]{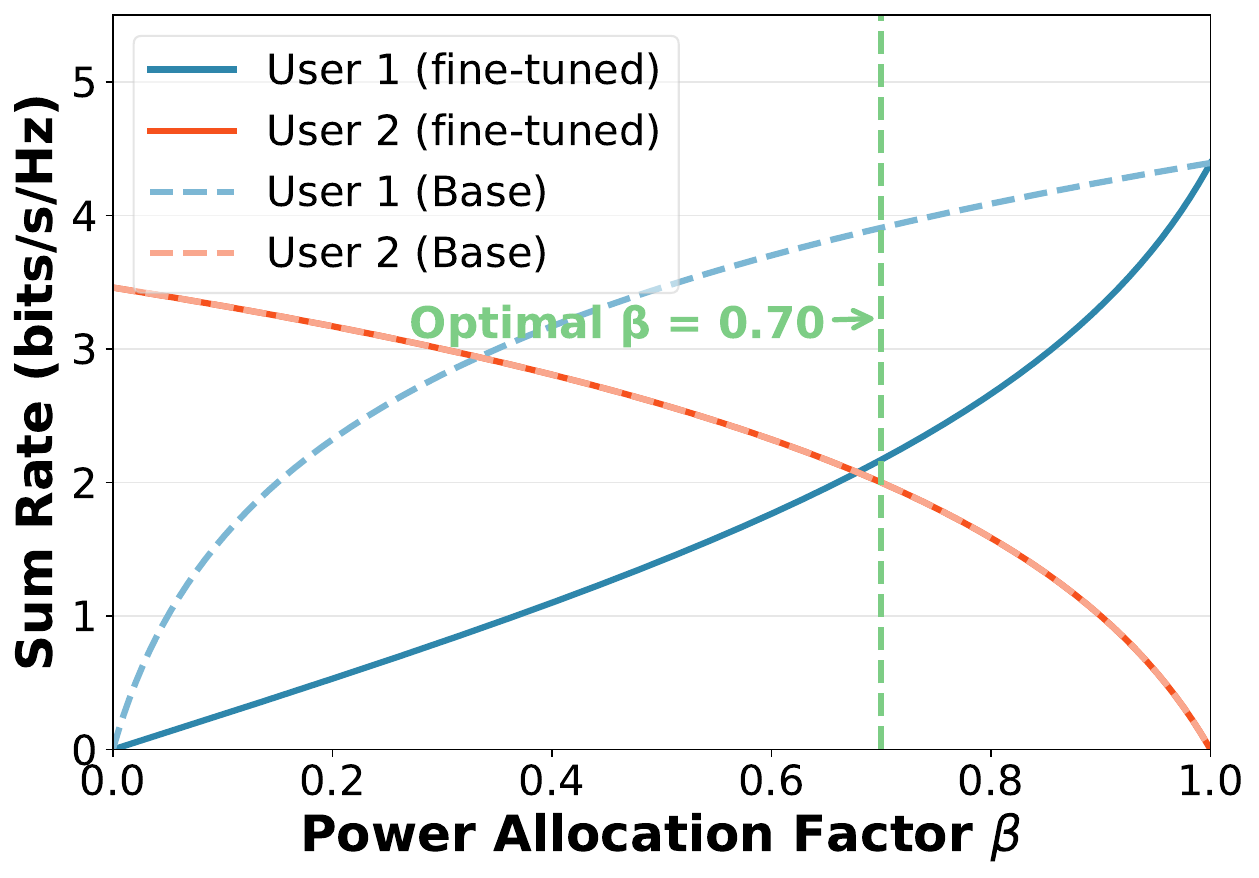}
        \vspace{-5mm}
        \caption{Power allocation problem.}
        \label{fig:noma_rate}
    \end{subfigure}
    \hfill
    \begin{subfigure}[htp]{0.49\linewidth}
        \centering
        \includegraphics[width=1\linewidth]{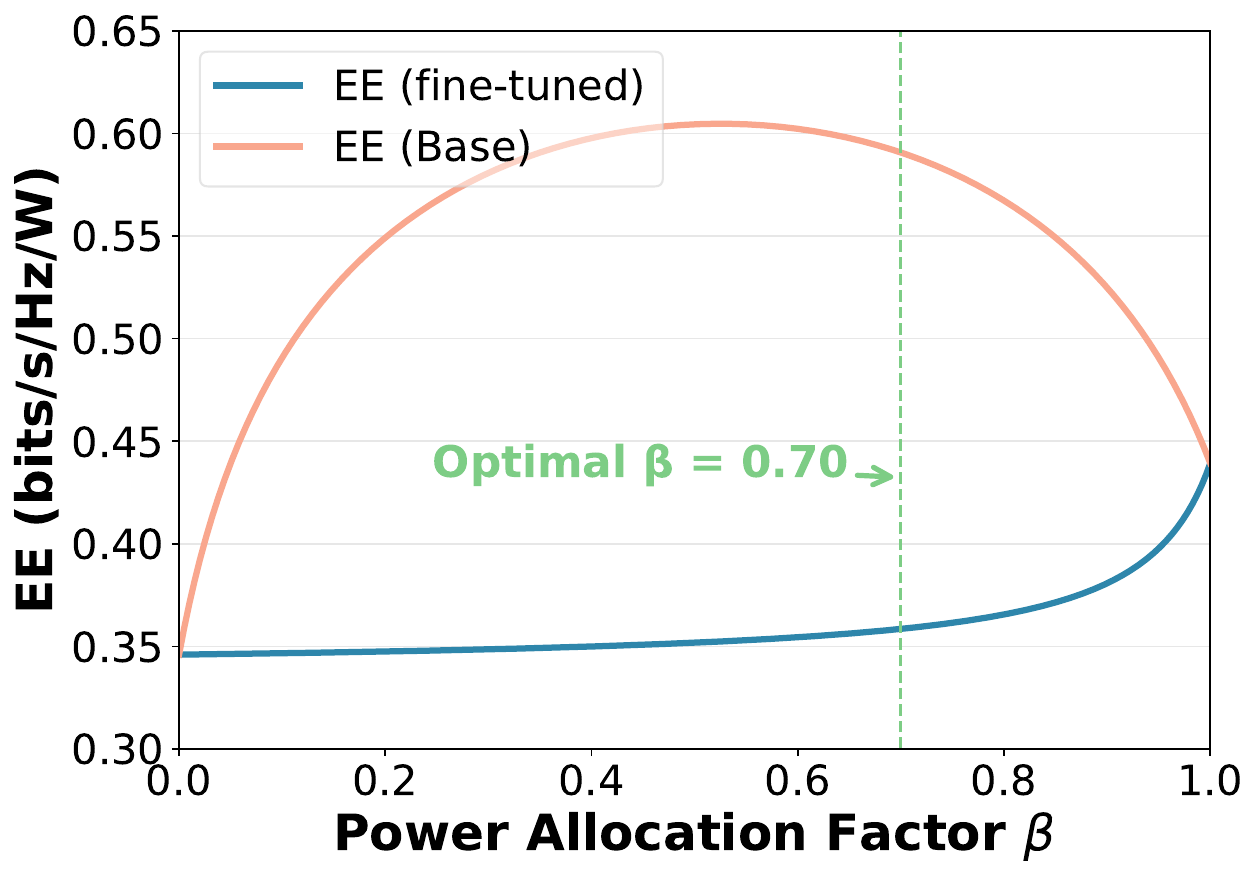}
        \vspace{-5mm}
        \caption{Energy efficiency problem.}
        \label{fig:noma_energy}
    \end{subfigure}
    \hfill
    \begin{subfigure}[htp]{0.49\linewidth}
        \centering
        \includegraphics[width=0.99\linewidth]{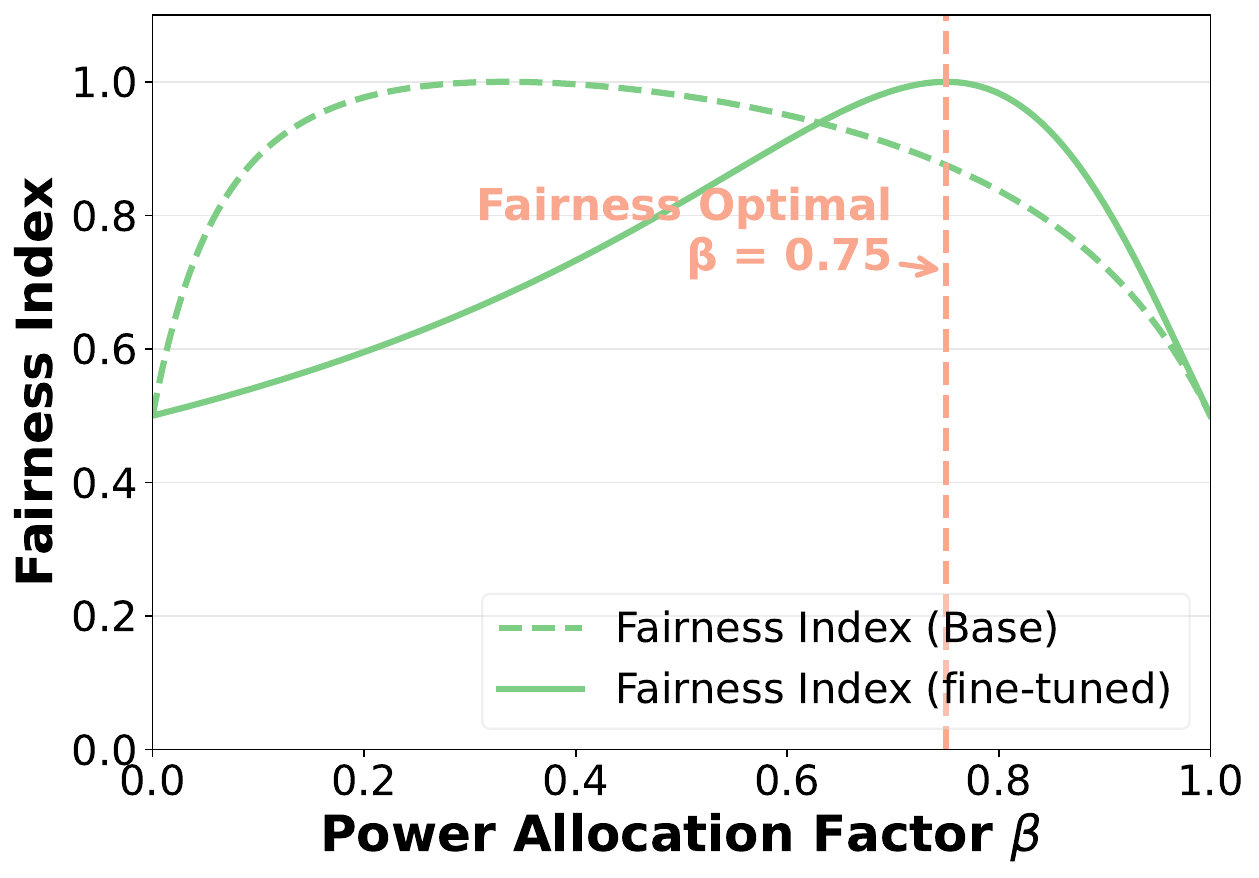}
        \vspace{-5mm}
        \caption{Fairness problem.}
        \label{fig:noma_fairness}
    \end{subfigure}
    \hfill
    \begin{subfigure}[htp]{0.49\linewidth}
        \centering
        \includegraphics[width=0.99\linewidth]{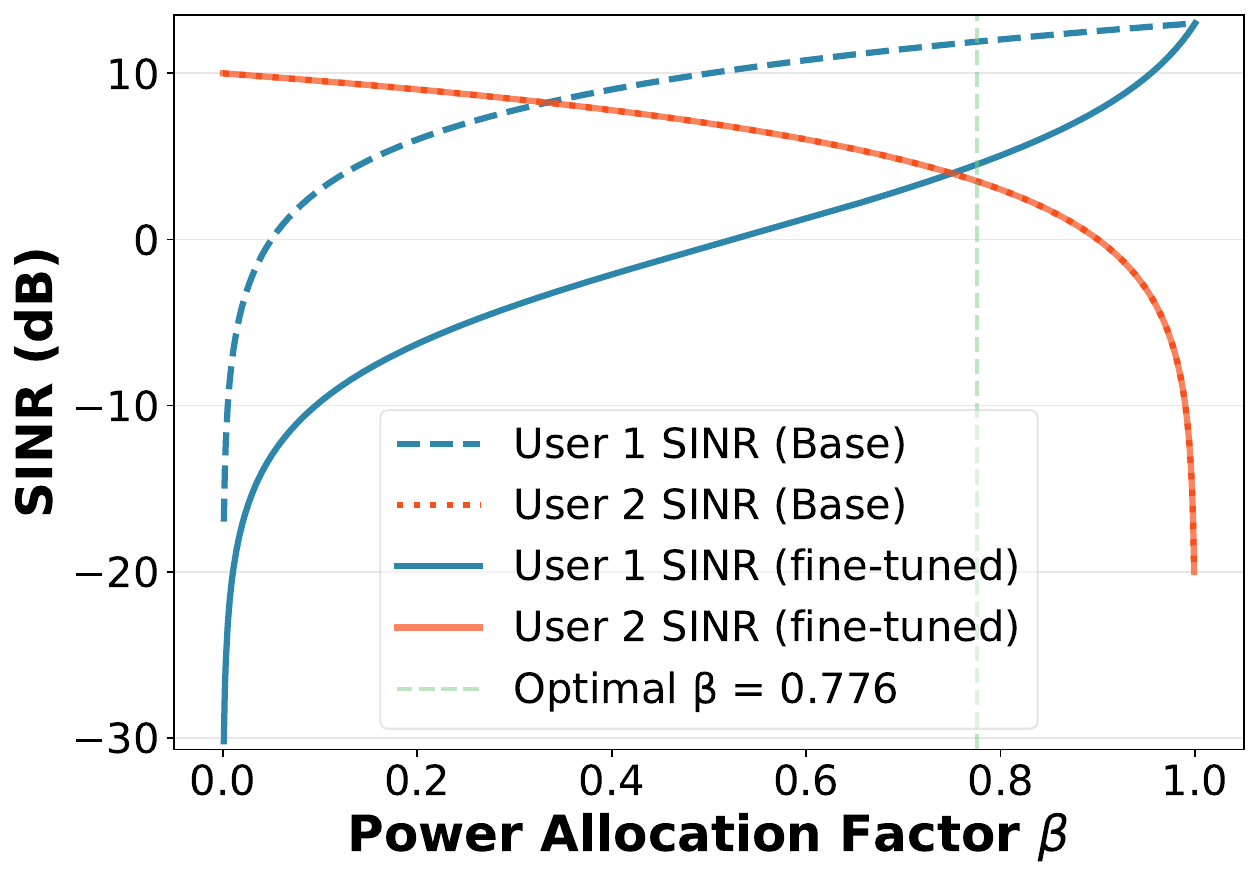}
        \vspace{-5mm}
        \caption{QoS problem.}
        \label{fig:noma_qos}
    \end{subfigure}
    \vspace{-2mm}
    \caption{Studies on the power allocation, energy efficiency, fairness and QoS in two-user NOMA case, using LLaMA-2 7B with and without fine-tuning with $R_{min} = 2$.}
    \label{fig:noma_problems}
\end{figure}

To further explore the potential and provide practical examples, we consider 4 two-user NOMA scenarios on sub-channel $i$, as briefly described in Section \ref{sec:III}. We test various prompt strategies, ultimately inputting the question in JSON format to enhance the LLaMA models' adherence to instructions. Although the NOMA case is relatively simple due to the limitations of the LLaMA 7B model in tackling complex problems, understanding and solving optimization problems remains crucial.
In this scenario, a quality of service (QoS) constraint is imposed on the second user, ensuring a minimum required data rate $R_{\min}$, i.e., $R_{i,2}^{\textrm{noma}}(\beta) \geq R_{\min}$. 

In Fig. \ref{fig:noma_problems}, the result demonstrates a significant discrepancy between the base model and the fine-tuned model in the NOMA power allocation problem. Specifically, the fine-tuned LLaMA 7b model identifies an optimal $\beta$ that maximizes the sum rate while satisfying the minimum rate requirement for user 2, while the base model fails to provide the correct simulation due to the incorrect formulation of the NOMA equations.
These findings highlight the significance of leveraging fine-tuned models with domain-specific expertise to ensure the reliability and accuracy of wireless communication simulations. 
Consequently, our study emphasizes the necessity of fine-tuning in developing and simulating advanced communication technologies. 

\subsection{Additional Evaluations}
We conduct additional experiments to further evaluate the fine-tuned models on two tasks: summarization and solving mathematical problems. For this purpose, we collect 200 wireless optimization problems and develop mathematical problems related to wireless communications to gain deeper insights and perform a more comprehensive evaluation (detailed in Appendix D).

Fig. \ref{fig:rouge_acc}(a) compares the summarization performance of the fine-tuned GPT-2 XL and LLaMA-2 7B models against baselines, using the ROUGE-1, ROUGE-2, and ROUGE-L metrics. The results, drawn from summarizing 200 published wireless optimization problems, demonstrate that fine-tuning substantially enhances the quality of the generated summaries. Notably, the fine-tuned LLaMA-2 7B model achieves pronounced improvements (+0.209 for both ROUGE-1 and ROUGE-L, and +0.193 for ROUGE-2), thereby surpassing the fine-tuned GPT-2 XL model and both base models. These findings underscore the effectiveness of domain-specific fine-tuning to generate more coherent, concise, and accurate summaries in complex technical fields such as wireless communications.
\begin{figure}[htp]
    \centering
    \includegraphics[width=1\linewidth]{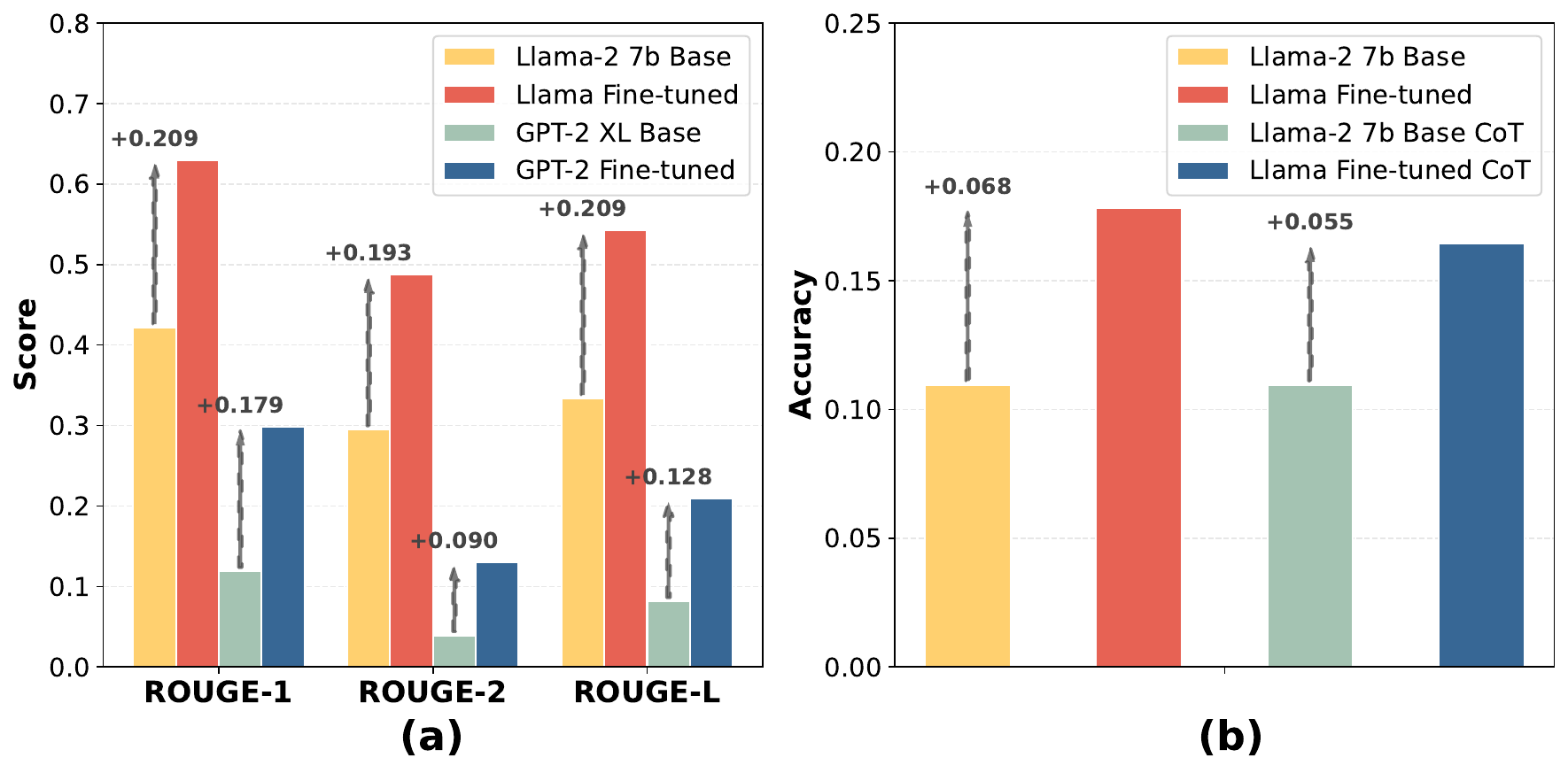}
    \caption{Performance gain comparisons for GPT-2 XL and LLaMA-2 7B models across summarization and mathematical problem-solving tasks.}
    \label{fig:rouge_acc}
    \vspace{-3mm}
\end{figure}

Fig. \ref{fig:rouge_acc}(b) highlights the impact of fine-tuning LLaMA-2 7B models with wireless-related context on their ability to solve mathematical problems. Although the gains are more modest than those observed in summarization, fine-tuning still yields accuracy improvements (+0.068 with CoT reasoning and +0.055 without CoT). Interestingly, the direct influence of CoT alone remains limited, suggesting that integrating domain-specific training data contributes more meaningfully to improved model performance than the introduction of reasoning prompts. This indicates a promising avenue for future research, where more targeted fine-tuning strategies with specialized mathematical datasets on wireless communications, potentially combined with advanced reasoning techniques, could further improve model accuracy and reliability in the domain.

\begin{figure}[h]
    \vspace{-2mm}
    \centering
    \includegraphics[width=0.95\linewidth]{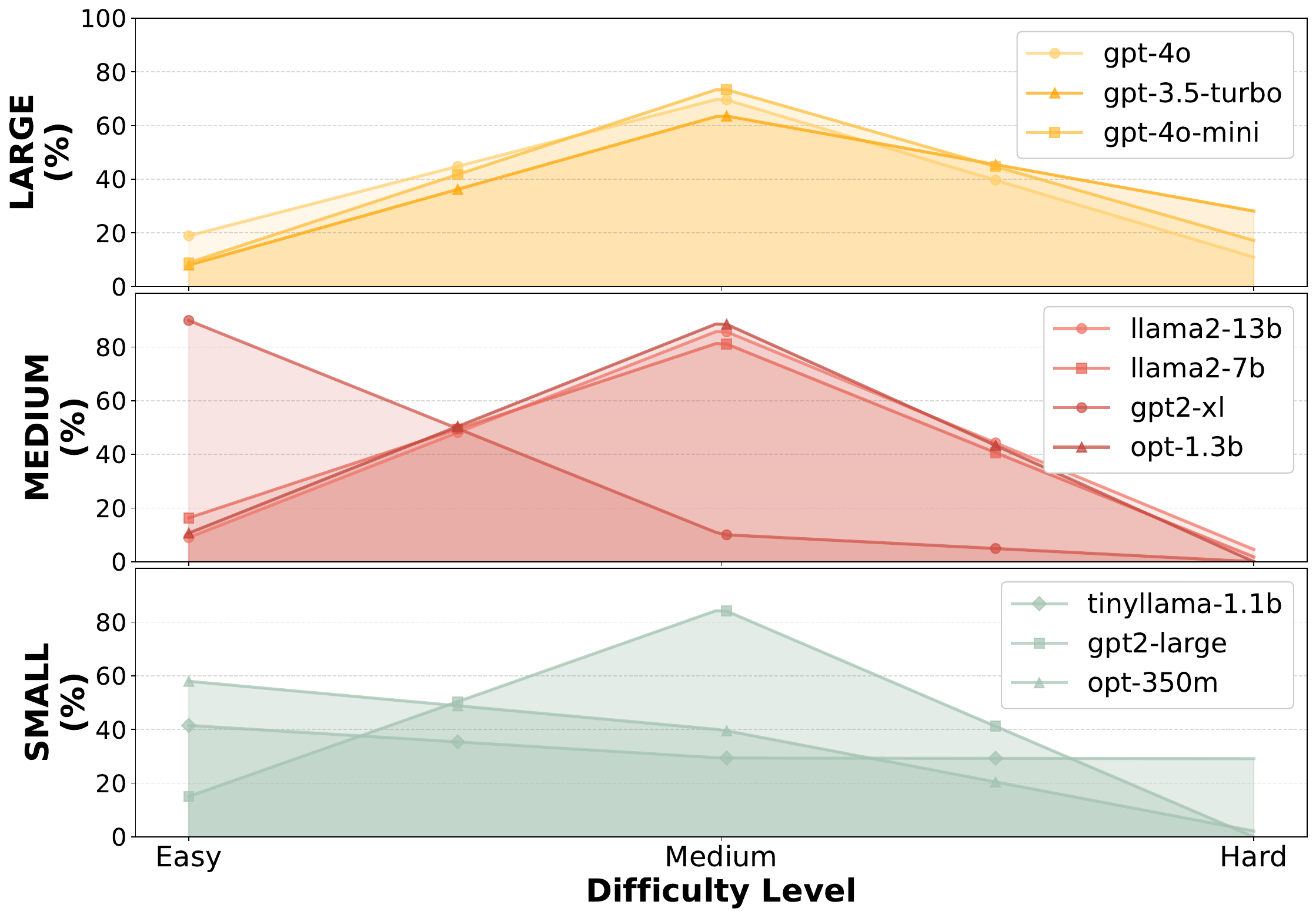}
    \caption{Comparison of difficulty distribution across LLMs.}
    \label{fig:difficulty}
    \vspace{-2mm}
\end{figure}

We also investigate how the models perform to determine the difficulty of the dataset in Fig. \ref{fig:difficulty}. Interestingly, the results show how models of varying sizes—large, medium, and small—assess question difficulty in the wireless communication dataset. Large models from the GPT-4 family show a balanced ability to classify questions across difficulty levels, with a peak at medium difficulty and some capacity to recognize harder questions. Medium-sized models (LLaMA2 and GPT2-XL) show a sharp decline in performance on hard questions, indicating limitations in handling more complex queries in this domain. Small models (GPT2-large) overwhelmingly label questions as easy or medium, with almost no questions categorized as hard, suggesting an underestimation of complexity in the dataset's context. These observations highlight that, while larger models are better suited for understanding wireless communication tasks, their substantial computational demands and impracticality for local deployments limit their applicability. To address these limitations, our proposed approach, which integrates a specialized dataset and a fine-tuning strategy—offers an effective solution to enhance the performance of smaller and more practical language models.

\vspace{-0.1em}
\section{Discussion}\label{sec:scaling}
In this section, we provide more insights into scaling laws for fine-tuning LLMs and the challenges these models face in wireless communication.

\vspace{-3mm}
\subsection{Scaling Laws}

Scaling laws describe the empirical relationships between a model’s parameter count, the volume of training data, and the computational resources utilized during fine-tuning. These relationships collectively determine the performance of the resulting model. Research indicates that increasing any of these factors typically leads to better performance \cite{fine_tune_literature3}. However, this improvement exhibits diminishing returns, ultimately reaching a performance ceiling where additional scaling yields minimal gains in fine-tuning tasks.
Scaling laws offer critical insights for guiding the training and fine-tuning strategies of LLMs \cite{fine_tune_literature4}. The existing literature on scaling laws has concentrated on various fields, e.g. biomedical and computer science \cite{fine_tune_literature1, fine_tune_literature2,fine_tune_literature4}, but there remains a notable gap in evaluating how scaling laws apply to fine-tuning in the field of wireless communication. To the best of our knowledge, there has been limited research on the simulations of scaling laws and the challenges associated with applying LLMs to wireless communications.

\begin{figure}[htp]
    \centering
    \vspace{-1mm}
    \includegraphics[width=0.9\linewidth]{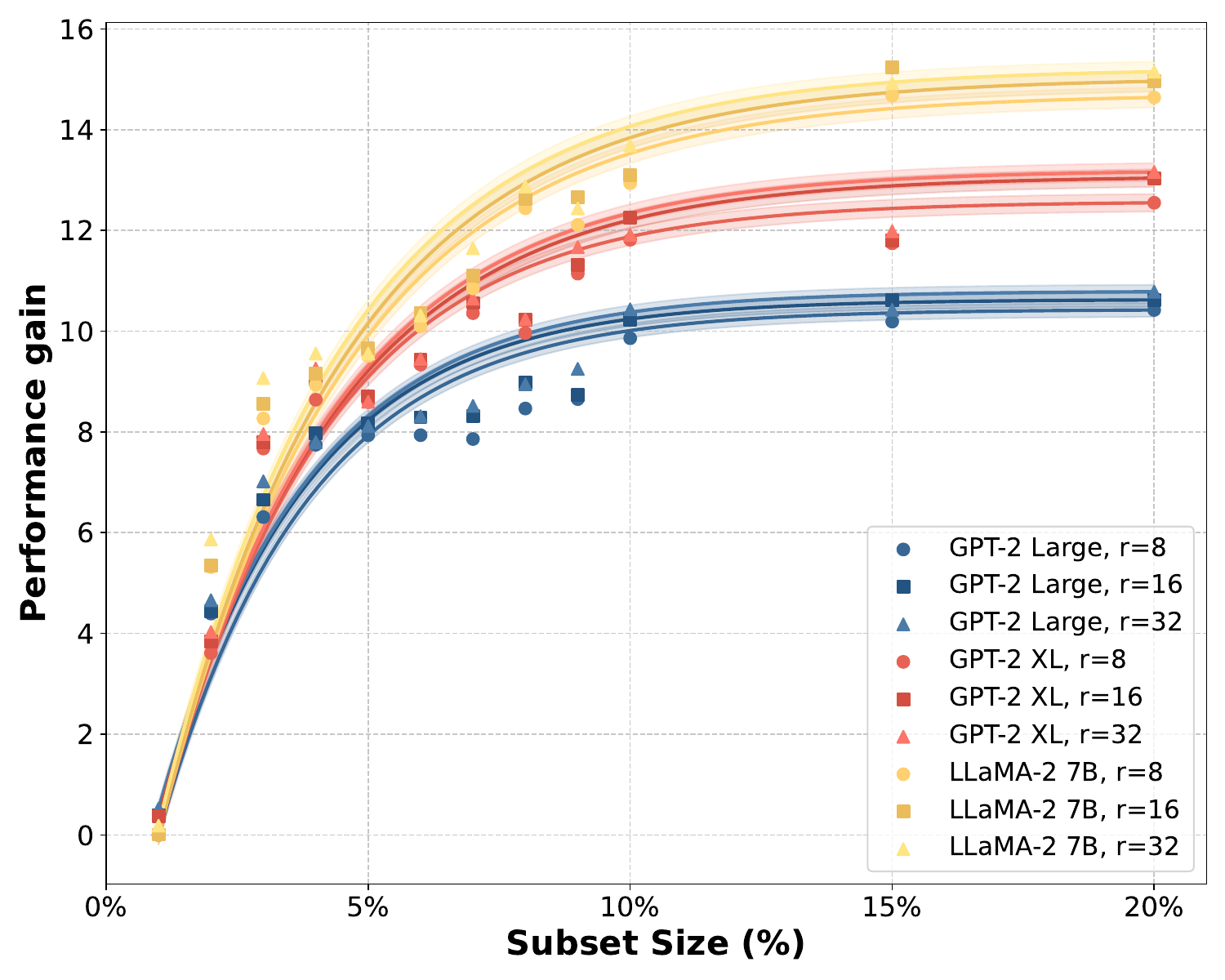}
    \caption{The performance scaling behavior of three different language models (GPT-2 Large, GPT-2 XL, and LLaMA-2 7B) across different subset sizes of training data.}
    \label{fig:scaling_alpha}
    \vspace{-2mm}
\end{figure}

We investigate the impact of rank $r$ on performance gain in LoRA for wireless communications tasks, as demonstrated in Section \ref{sec:IV}.
As illustrated in Fig. \ref{fig:scaling_alpha}, while increasing the volume of wireless data generally enhances the performance of the model, it eventually approaches an inherent ceiling. This observation highlights that merely increasing data quantity is insufficient for sustained improvements. Instead, optimal fine-tuning requires high-quality data to ensure precise task alignment and effective training within the constraints of the performance ceiling.
The result indicates that while all models benefit from more training data, the returns diminish after about 10-15\% of the data. In addition, the performance gain for the different sizes of the models is limited when the value of $r$ increases. In practice, it also suggests that the value of $r$ can be set relatively small for computationally limited devices to balance computational load and performance.

\vspace{-0.2em}
\subsection{Challenges for LLMs in Wireless Communications}

\begin{table}[ht]
\centering
\setlength{\tabcolsep}{4pt} 
\renewcommand{\arraystretch}{1.2} 
\vspace{-2mm}
\caption{Model performance across subject areas.}
\begin{tabular}{lcccc}
\toprule
\textbf{Model} & \textbf{Marketing} & \textbf{Social Science} & \textbf{History} & \textbf{Wireless Comm.} \\
\midrule
LLaMA 2 7B          & $0.284$ & $0.271$ & $0.210$ & $0.121\quad \textcolor{red}{\downarrow}$ \\
LLaMA 2 13B         & $0.306$ & $0.287$ & $0.233$ & $0.149\quad \textcolor{red}{\downarrow}$ \\
GPT-2 XL            & $0.406$ & $0.374$ & $0.322$ & $0.325\quad \textcolor{red}{\phantom{\downarrow}}$ \\
GPT-3.5 turbo       & $0.612$ & $0.603$ & $0.457$ & $0.364\quad \textcolor{red}{\downarrow}$ \\
GPT-4o-mini         & $0.840$ & $0.831$ & $0.669$ & $0.640\quad \textcolor{red}{\downarrow}$ \\
GPT-4o              & $0.841$ & $0.864$ & $0.701$ & $0.651\quad \textcolor{red}{\downarrow}$ \\
\bottomrule
\end{tabular}
\label{tab:wireless_performance}
\end{table}

In Table \ref{tab:wireless_performance}, we provide the performance of various LLMs across four domains: marketing, social science, history, and wireless communications. The datasets for marketing, social science, and history are generated using the proposed framework detailed in Section \ref{sec:III}. The results indicate that while models like GPT-4o achieve high accuracy in marketing (0.841), social science (0.864), and history (0.701), their performance decreases in wireless communications, with scores ranging from 0.121 for LLaMA 2 7B to 0.651 for GPT-4o. This consistent underperformance in the wireless communications domain highlights the challenges LLMs face with highly technical and specialized content, likely due to limited domain-specific knowledge and complex terminology. Moreover, increasing model size does not sufficiently bridge this performance gap, suggesting that targeted fine-tuning or the integration of specialized knowledge bases is necessary to enhance LLM efficacy in specialized fields.

In this light, our refined dataset generation and fine-tuning strategy aims to provide a clear and effective framework for understanding and optimizing LLMs. This framework aims to leverage domain-specific data and customized training methods to enhance the adaptability, accuracy, and interpretability of LLMs in wireless communication applications. By systematically addressing the unique challenges of the field, our approach facilitates the development of more reliable and efficient models, thereby advancing the practical deployment of LLMs.

\section{Conclusion} \label{sec:VII}
This work introduces a domain-specific dataset designed to allow LLMs to address complex tasks in wireless communications by incorporating multi-hop reasoning and employing a theoretically justified PVI-based fine-tuning methodology. Our experimental validation on multiple benchmarks demonstrates the effectiveness of our approach, highlights notable improvements over existing methods, and establishes a strong foundation for the integration of LLMs into wireless networks. Future research could focus on expanding the dataset to include optimization problems and fostering advanced techniques such as convex optimization and machine learning-based methods. Additionally, enhancing multi-hop question generation tailored to interconnected concepts in wireless communications may further improve LLM reasoning capabilities, driving innovations in smarter, more efficient network designs.

\begin{figure*}[t!]
    \vspace{-2mm}
    \centering
    \begin{subfigure}[t]{0.32\textwidth}
        \centering
        \includegraphics[height=4.5cm, keepaspectratio]{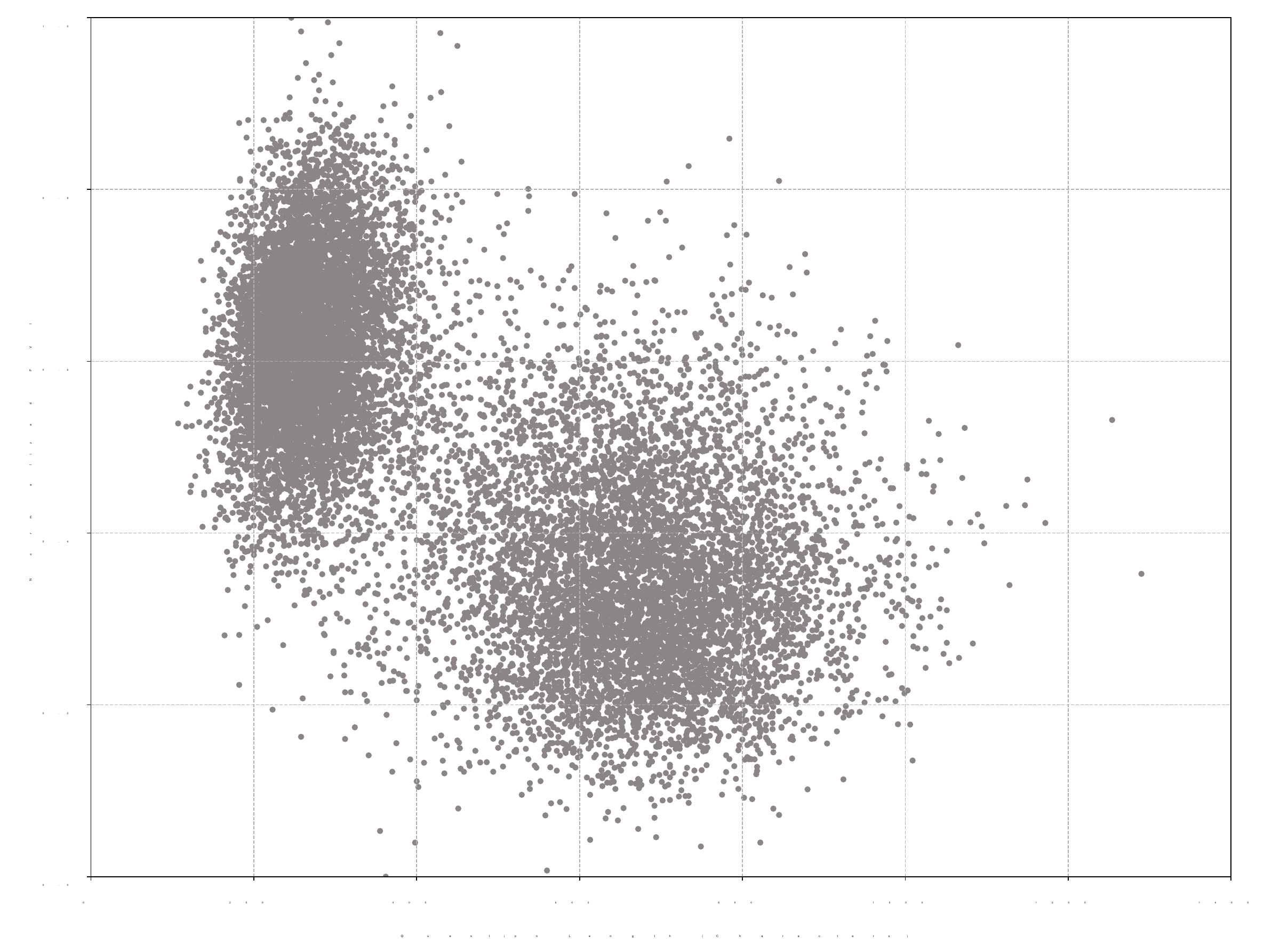}
        \caption{Question length with difficulty level.}
        \label{fig:length_pvi}
    \end{subfigure}
    \hfill
    \begin{subfigure}[t]{0.32\textwidth}
        \centering
        \includegraphics[height=4.5cm, keepaspectratio]{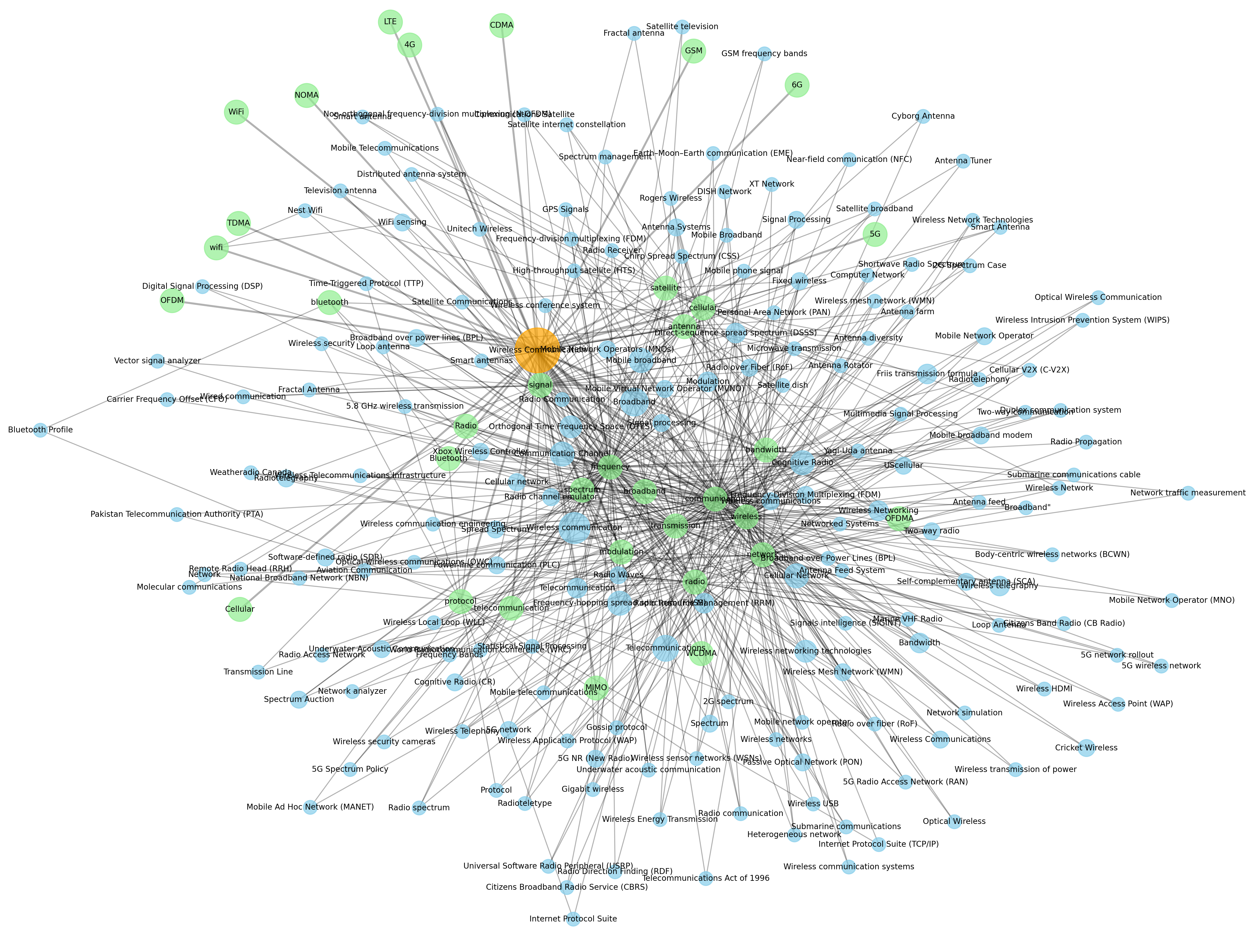}
        \caption{Graph relations of part of the entities.}
        \label{fig:wireless_graph}
    \end{subfigure}
    \hfill
    \begin{subfigure}[t]{0.32\textwidth}
        \centering
        \includegraphics[height=4.5cm, keepaspectratio]{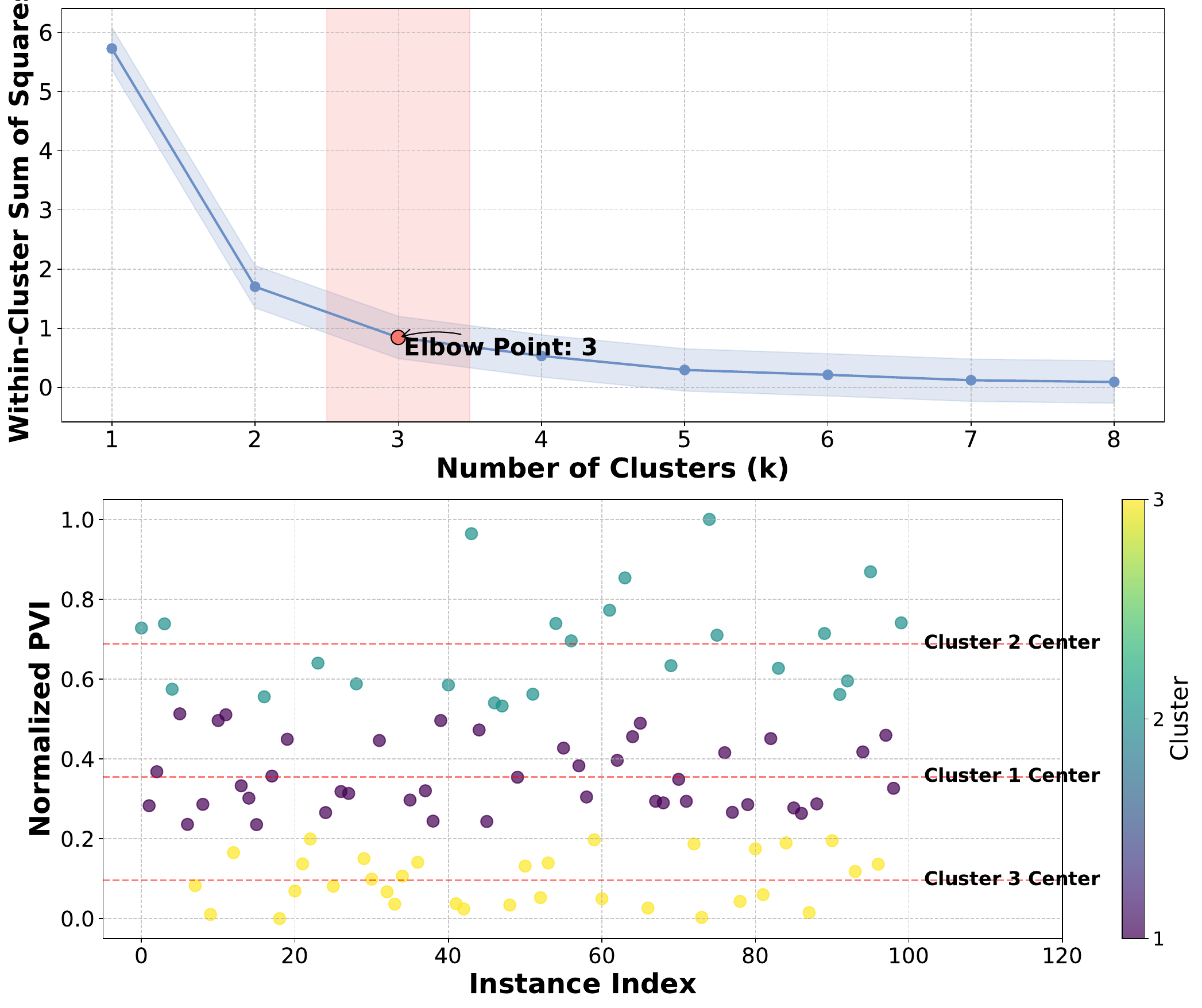}
        \caption{K-means clustering analysis of PVI.}
        \label{fig:kmeans_clustering_plot}
    \end{subfigure}
    \caption{
        (a) Question length with difficulty level. Each point represents a single question with the corresponding question length and its normalized difficulty level (PVI).
        (b) Graph relations of entities in wireless communication. 35\% of the entities are randomly selected to avoid overcrowding.
        (c) K-means clustering analysis of PVI values. The top plot shows the elbow method to determine the optimal number of clusters \cite{Lin_elbow,cui_elbow}. The bottom plot clusters normalized PVI values into three groups.
    }
    \label{fig:combined_figure}
\end{figure*}

\section*{Appendix A: Additional Simulations}\label{Appendix_A}

We provide an analysis of question length and difficulty levels in Fig. \ref{fig:combined_figure}\textcolor{blue}{(a)}, graph relations of part of the entities extracted in Fig. \ref{fig:combined_figure}\textcolor{blue}{(b)}, and K-means clustering analysis of PVI values of 100 randomly selected questions in Fig. \ref{fig:combined_figure}\textcolor{blue}{(c)}.

\section*{Appendix B: Example of Generated Questions}\label{Appendix_B}

\begin{figure}[htp]
    \vspace{-3mm}
    \centering
    \includegraphics[width=1\linewidth]{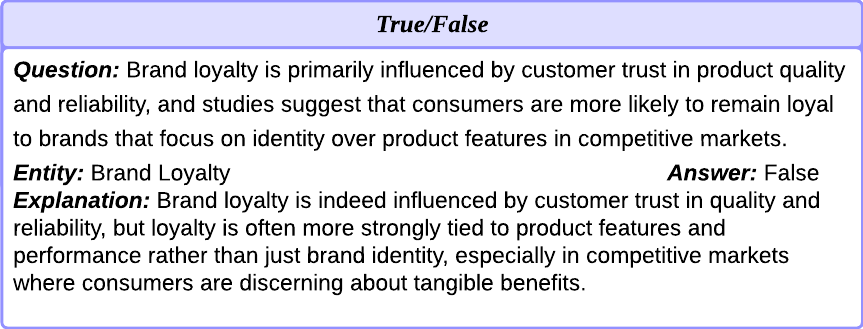}
    \caption{Example question in the subject marketing.}
    \vspace{-3mm}
    \label{fig:example_area1}
\end{figure}

We list one of the examples of the generated questions for the other subject demonstrated in Section \ref{sec:III}. Due to space limitations, we only show the true/false questions generated for the field of marketing in Fig. \ref{fig:example_area1} to demonstrate the results in Table \ref{tab:wireless_performance}. 2,000 multi-hop questions are generated for each subject area, including multiple-choice and true/false questions. For wireless communication-related questions, 2,000 questions are randomly selected from the generated dataset.  

\section*{Appendix C: Prompts}\label{Appendix_C}
We present the prompts used for data generation, evaluation, and curation in this study. The organization of prompts for each language model differs slightly. For example, when evaluating datasets with GPT-3.5 Turbo, GPT-4o-mini, and GPT-4o via the OpenAI API, the “system” and “user” prompts must be specified separately. However, the underlying conceptual framework for prompt remains consistent across models.

The Quiet-STaR \cite{Quiet-STaR} introduces a novel token-by-token sampling algorithm designed to optimize context representation during entity extraction and question integration. Although Quiet-STaR does not mandate a specific prompt template, the prompts utilized in this study, are designed to identify and mitigate biases that may arise the integration of questions derived from multiple entities. These prompts are constructed on a basis of the version of the methodological objectives of this paper. By leveraging this algorithm, LLMs autonomously enhance the integration of questions, leading to improvements in both coherence and representational accuracy.

To evaluate and address potential biases in the generated prompts, we define bias as follows.
\begin{figure}[H]
        \vspace{-2mm}
        \centering
        \includegraphics[width=1\linewidth]{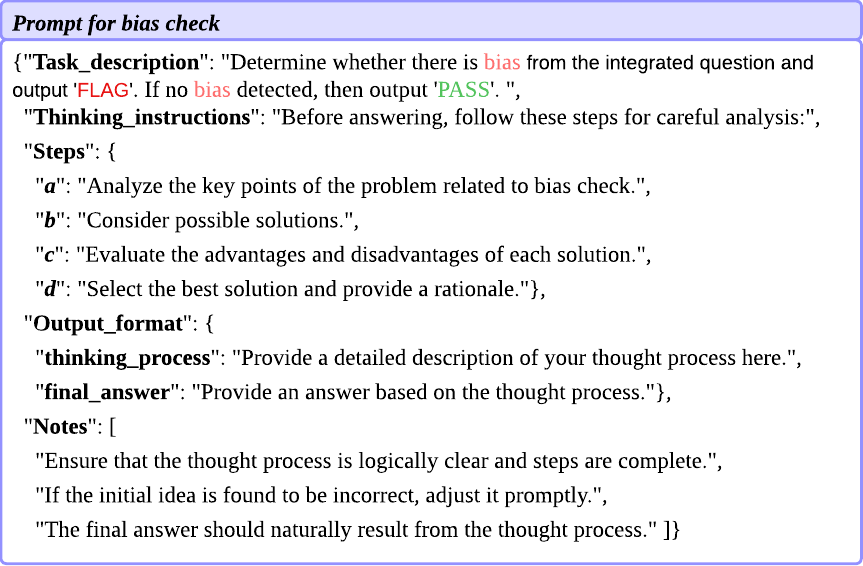}
        \caption{Prompt of determining the potential bias.}
        \label{fig:prompts_bias}
\end{figure}

\begin{enumerate}
    \item \textbf{Selection Bias}: Occurs when the integrated question disproportionately incorporates subquestions from specific paragraphs or topics, leading to an over-representation of certain contextual aspects while neglecting others.
    \item \textbf{Contextual Bias}: Refers to the incorporation of assumptions about the subject matter into the integrated question, which may distort or misrepresent the information extracted from the source material.
    \item \textbf{Order Bias}: Arises when the integration process prioritizes subquestions based on their order of appearance, thereby influencing the weighting of information from earlier or prominently positioned content.
\end{enumerate}

\begin{figure}[htp]
    \centering
    \vspace{-2mm}
    \begin{subfigure}[htp]{0.99\linewidth}
        \centering
        \includegraphics[width=\linewidth]{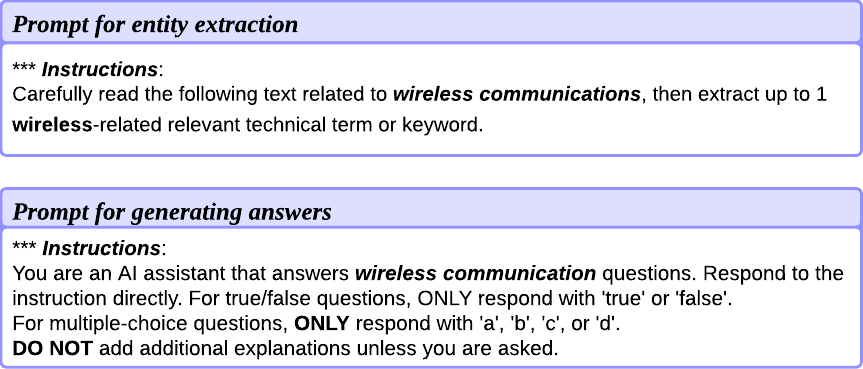}
        \caption{Prompts of entity extraction and answer generation.}
        \label{fig:prompts}
    \end{subfigure}
    \hfill
    \begin{subfigure}[htp]{1\linewidth}
        \centering
        \includegraphics[width=\linewidth]{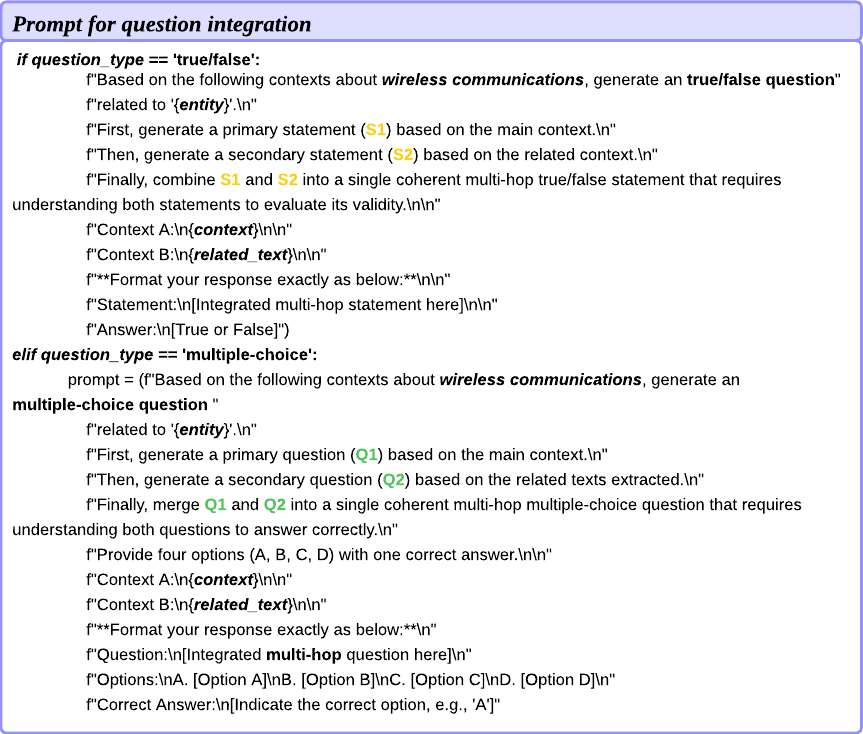}
        \caption{Prompt for question integration. }
        \label{fig:prompt1}
    \end{subfigure}
    \caption{Prompts of entity extraction, question answering, and question integration in Section \ref{sec:III}.}
    \vspace{-3mm}
    \label{fig:combined_prompts}
\end{figure}

\section*{Appendix D: Multi-Agents for Question Generation}
Multi-agent systems for LLMs have been extensively explored in recent research \cite{multiagent1}. These systems involve multiple LLMs working together to perform tasks such as reasoning, problem solving, and decision-making. The potential benefits of multi-agent systems include improved efficiency, innovation, and adaptability \cite{multiagent}. However, there is limited literature on generating synthetic datasets using the multi-agent framework.
To create mathematical problems in wireless communications, we explore the methodology of generating synthetic maths problems by utilizing the proposed multi-agent framework. A concise summary of how the considered multi-agents interact to generate and refine mathematical problems related to NOMA systems is provided.

In the proposed framework, six specialized agents—\textbf{Solvix, ProbMaster, PrimeArchitect, Validata, RefineMaster, and ExploreEnhancer}-collaborate to produce high-quality NOMA-related mathematical problems. Each agent focuses on a specific stage of content generation, validation, or enhancement, ensuring that the final output is both mathematically rigorous and instructionally valuable.

\begin{enumerate} 
\item \textbf{Solvix}: Generates accurate, step-by-step solutions for NOMA-related mathematical problems. 
\item \textbf{ProbMaster}: Creates clear statements of NOMA-aligned mathematical problems from instructions or existing solutions. 
\item \textbf{PrimeArchitect}: Develops diverse NOMA problems covering topics like power allocation and SINR calculations. 
\item \textbf{Validata}: Validates problems and solutions for accuracy, adherence to NOMA principles, and clarity, providing detailed feedback to identify and rectify any inconsistencies or errors.
\item \textbf{RefineMaster}: Enhances problem statements and solutions based on Validata’s feedback, ensuring they remain challenging and educational. 
\item \textbf{ExploreEnhancer}: Integrates advanced NOMA concepts into problems, increasing the diversity of questions without compromising solvability. \end{enumerate}

The workflow can be summarized as follows.
The generation process begins with either direct question generation (\textbf{PrimeArchitect} or \textbf{ProbMaster}) or solution-first approach (\textbf{Solvix}, then \textbf{PrimeArchitect}).
For direct question generation, the agent produces a mathematical problem statement aligned with the NOMA principles. This approach is straightforward but can lead to overly complex or incoherent questions that the model struggles to solve. The solution-first approach is for more challenging problems. The process starts with having \textbf{Solvix} produce a detailed, step-by-step solution. Once a coherent and correct solution is established, \textbf{PrimeArchitect} reverse-engineers the question from that solution. This ensures that the final problem is solvable, coherent, and aligned with NOMA-related constraints, improving the validity of the generated content.
After a preliminary problem (or its derived solution if applicable), \textbf{ExploreEnhancer} introduces advanced NOMA concepts. For instance, incorporating additional dimensions like imperfect SIC, user mobility, multi-cell interference, etc. Once a stable and validated problem and solution pair is obtained, \textbf{Validata} rigorously evaluates it for mathematical accuracy, NOMA principle adherence, and consistency.

\begin{figure}[htp]
    \centering
    \begin{subfigure}[b]{0.99\linewidth}
        \centering
        \includegraphics[width=\linewidth]{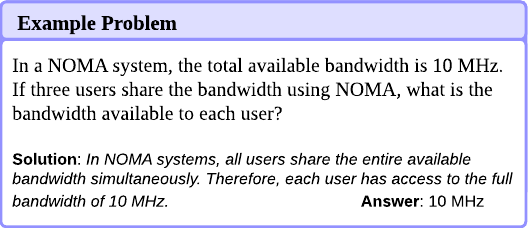}
        \caption{Example question 2.}
        \label{fig:q1}
    \end{subfigure}
    \hfill
    \begin{subfigure}[b]{0.99\linewidth}
        \centering
        \includegraphics[width=\linewidth]{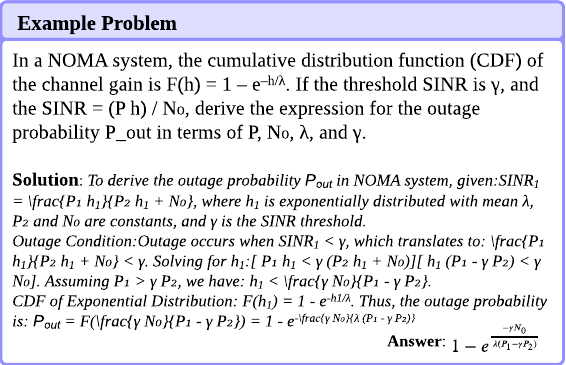}
        \caption{Example question 3.}
        \label{fig:q3}
    \end{subfigure}
    \caption{Combined figure showing Example questions 2 and 3.}
    \label{fig:combined}
\end{figure}

To maintain the quality of the generated questions, we use GPT-4o-mini to filter out similar questions. Subsequently, domain experts review the questions, along with their corresponding explanations and answers, to further ensure the quality and mitigate hallucinations in LLMs. Ultimately, 73 questions are retained from an initial set of 200. We narrow the number of questions to 200 due to the high cost of the question generation, mainly because of the validation of the solutions for each instance and the communication between agents. For detailed prompts, please refer to the code on GitHub. In Fig. \ref{fig:combined}\textcolor{blue}{(b)}, we list more examples of generated questions. 

It is noteworthy that type 2 example questions were evaluated using various state-of-the-art models. Some large models, including Falcon-40B, failed to produce correct answers, instead returning responses such as 3.33MHz. This indicates a lack of meaningful understanding of certain concepts in wireless communications. Additionally, although some advanced models, such as LLaMA 3.1 70B, initially provide correct answers, they tend to deliver incorrect responses when queried again using an interrogative tone.
\FloatBarrier
\bibliographystyle{IEEEtran}
\bibliography{WirelessDataset}
\end{document}